\documentclass[sigconf,nonacm]{acmart}

\AtBeginDocument{%
  \providecommand\BibTeX{{%
    \normalfont B\kern-0.5em{\scshape i\kern-0.25em b}\kern-0.8em\TeX}}}

\setcopyright{acmcopyright}
\acmPrice{15.00}
\acmDOI{10.1145/3540250.3549120}
\acmYear{2022}
\copyrightyear{2022}
\acmSubmissionID{fse22main-p427-p}
\acmISBN{978-1-4503-9413-0/22/11}
\acmConference[ESEC/FSE '22]{Proceedings of the 30th ACM Joint European Software Engineering Conference and Symposium on the Foundations of Software Engineering}{November 14--18, 2022}{Singapore, Singapore}
\acmBooktitle{Proceedings of the 30th ACM Joint European Software Engineering Conference and Symposium on the Foundations of Software Engineering (ESEC/FSE '22), November 14--18, 2022, Singapore, Singapore}



\usepackage{url}            % simple URL typesetting
\usepackage{booktabs}       % professional-quality tables
\usepackage{amsfonts}       % blackboard math symbols
\usepackage{nicefrac}       % compact symbols for 1/2, etc.
\usepackage{microtype}      % microtypography

\usepackage{multirow}

\usepackage{enumitem}
\usepackage[utf8]{inputenc}

\usepackage{listings}
\usepackage{caption}
\usepackage{pifont}
\usepackage{amsmath}
\usepackage{microtype}
\usepackage{tabularx}
\usepackage{verbatim}
\usepackage{capt-of}
\usepackage{paracol}
\usepackage{fancyvrb}
\usepackage{bm}
\usepackage{pifont}
\usepackage{longtable}
\usepackage{svg}
\usepackage[frozencache]{minted} %arxiv
\usepackage{xspace}

\graphicspath{ {./img/} }

\usepackage{comment}
\usepackage[title]{appendix}
\usepackage{prettyref}%
\newrefformat{sec}{Sect.~\ref{#1}}%
\newrefformat{def}{Def.~\ref{#1}}%
\newrefformat{fig}{Fig.~\ref{#1}}%
\newrefformat{tab}{Tab.~\ref{#1}}%
\newrefformat{alg}{Algorithm~\ref{#1}}%
\newrefformat{fun}{Function~\ref{#1}}%
\newrefformat{lst}{Listing~\ref{#1}}%
\newrefformat{appendix}{Appendix~\ref{#1}}%

\newcommand{\Real}{\ensuremath{\mathbb{R}}}
\newcommand{\Int}{\ensuremath{\mathbb{Z}}}
\DeclareMathOperator*{\argmin}{arg\,min}

\DeclareMathOperator{\relu}{ReLU}

\newcommand{\ourTool}{\texttt{\ensuremath{\nu}Term}\xspace}

\definecolor{pblue}{rgb}{0.13,0.13,1}
\definecolor{pgreen}{rgb}{0,0.5,0}
\definecolor{pred}{rgb}{0.9,0,0}
\definecolor{pgrey}{rgb}{0.46,0.45,0.48}

\usepackage{subcaption}

\usepackage{tikz}
\usetikzlibrary[positioning,calc]
\usepackage{pgfplots}
\usepackage{pgfplotstable}

\pgfplotsset{
  cycle list={
    {blue, mark=+},
    {red, mark=x},
    {brown, mark=square}
  },
  every tick label/.append style={font=\tiny},
  every axis legend/.append style={font=\scriptsize}
}

\usepackage{array}
\newcolumntype{L}[1]{>{\raggedright\let\newline\\\arraybackslash\hspace{0pt}}m{#1}}
\newcolumntype{C}[1]{>{\centering\let\newline\\\arraybackslash\hspace{0pt}}m{#1}}
\newcolumntype{R}[1]{>{\raggedleft\let\newline\\\arraybackslash\hspace{0pt}}m{#1}}

\newcounter{cnt:snote}
\makeatletter
\newcommand\snote{\stepcounter{cnt:snote}\@ifstar{\@snotemar}{\@snotepar}}
\newcommand{\@snotepar}[2]{\noindent \marginpar{\textbf{\color{orange}\arabic{cnt:snote}:} #2}
  {\color{orange}\bf (}#1{\color{orange} \bf )\textsuperscript{\arabic{cnt:snote}}}}
\newcommand{\@snotemar}[1]{\marginpar{\textbf{\arabic{cnt:snote}\\}
		\@snotebody{#1}}}
\newcommand{\@snotebody}[1]{\textsf{#1}}
\makeatother

\definecolor{ao}{rgb}{0.0, 0.5, 0.0}

\usepackage{listings}
\definecolor{lightgray}{gray}{0.9}

\begin{document}

\title{Neural Termination Analysis}

\author{Mirco Giacobbe}
\authornote{The authors are listed in
  alphabetical order regardless of individual contributions or seniority.}
%\email{m.giacobbe@bham.ac.uk}
\affiliation{%
  \institution{University of Birmingham}
  \country{UK}
}

\author{Daniel Kroening}
\authornote{This work was done prior to joining Amazon.}
\authornotemark[1]
%\email{daniel.kroening@magd.ox.ac.uk }
\affiliation{%
  \institution{Amazon, Inc.}
  \country{USA}
}

\author{Julian Parsert}
\authornotemark[1]
%\email{julian.parsert@stcatz.ox.ac.uk}
\affiliation{%
  \institution{University of Oxford}
  \country{UK}
}

\begin{abstract}
  We introduce a novel approach to the automated termination analysis of
computer programs: we use neural networks to represent ranking functions. 
Ranking functions map program states to values that are bounded from below
and decrease as a program runs; the existence of a ranking function proves
that the program terminates.  We train a neural network from sampled
execution traces of a program so that the network's output decreases along
the traces; then, we use symbolic reasoning to formally verify that it
generalises to all possible executions.  Upon the affirmative answer we obtain a
formal certificate of termination for the program, which we call a neural
ranking function.  We demonstrate that thanks to the ability of neural
networks to represent nonlinear functions our method succeeds over programs
that are beyond the reach of state-of-the-art tools.  This includes programs
that use disjunctions in their loop conditions and programs that include
nonlinear expressions.

\end{abstract}

\maketitle

\section{Introduction}\label{sec:intro}

Software is a complex artefact.  Programming is prone to error and some bugs
are hard to find even after extensive testing.  Bugs may cause crashes,
undesirable outputs, and can prevent a program from responding at all
which causes poor performance and can be a vulnerability~\cite{opensslbug}.   
Termination analysis addresses the question of whether, for every possible
input, a program halts.  This is undecidable in general, yet tools that work
in practice have been developed by industry and academia~\cite{CookPR06,
HeizmannCDEHLNSP13, BrockschmidtCIK16, GieslABEFFHOPSS17, LeAFKN20}.  In
this paper, we introduce a novel technique that effectively trains neural
networks to act as formal proofs of termination and, thanks to the
expressivity of neural networks, significantly extends the set of programs
that can be proven to terminate automatically.

To argue that a program terminates one usually presents a \emph{ranking
  function} for each loop in the program.  Ranking functions map program states
to values that (i)~decrease by a discrete amount after every loop iteration and
(ii)~are bounded from below~\cite{Floyd67}.  They are certificates of
termination: if a ranking function exists, then the program terminates for every
possible input.
\begin{figure}
  \begin{minted}{Java}
  int x, y, z;
  ...
  while (x < y || x < z) {
    x++;
  }
  \end{minted}
  \caption{A simple program with disjunctive loop guard.}
  \label{fig:disj}
  \end{figure}

Many existing methods find ranking functions by relying on symbolic
reasoning~\cite{ArtsG00, PodelskiR04, BradleyMS05, Cousot05, pldi/CookPR06,
ChenXYZZ07, DBLP:journals/fmsd/CookKRW13, icalp/BradleyMS05,
10.1007/978-3-642-54862-8_12, Urban13, 10.1007/978-3-642-36742-7_4,
GieslABEFFHOPSS17, BrockschmidtOEG10, rta/OttoBEG10, DBLP:conf/sas/Spoto16,
DBLP:conf/fmco/AlbertAGPZ07, HeizmannHP14, DBLP:conf/pldi/ChenHLLTTZ18}. 
Moreover, they typically focus on the case of linear ranking functions
for programs that can be represented as conjunctions of linear constraints. 
For this particular case, Farkas' lemma offers a means to compute the ranking
function efficiently~\cite{PodelskiR04}. However, finding proofs for programs
that use disjunctive (e.g., Fig.~\ref{fig:disj}) or nonlinear loop guards
is much more difficult. 

Our approach is based on the principle that \emph{finding a
proof is much harder than checking that a given candidate proof is valid}.
We use machine learning to guess a proof followed by symbolic reasoning to verify it:
first, we learn a candidate ranking function by training
a neural network that decreases along sampled runs of the program;
then, we use satisfiability modulo theories (SMT) solving to check whether
this candidate {\em neural ranking function} (NRF)
decreases along every possible run of the program.
We use networks that are always bounded from below,
thus upon success we have a proof of termination.

Our experiments with established termination benchmarks de\-monstrate that the
idea is effective: most loops can be proven to terminate using
very tiny neural networks (fewer than 10 neurons) and at most a thousand
sample runs.  We give exemplars of neural architectures and loss functions
for training monolithic and lexicographic ranking functions.  Using a neural
network with just one hidden layer and a straight-forward training routine, our
method discovers neural ranking functions for over 75\% of the benchmarks in
a standard problem set for termination
analysis~\cite{DBLP:conf/tacas/GieslRSWY19, 10.1007/978-3-030-45237-7_21}. 
Our method subsumes a broad range of existing
termination analysis strategies: not only do we discover linear ranking functions,
but also ranking functions for problems that require piecewise linear or
lexicographic termination arguments~\cite{HeizmannHP14,UrbanGK16}.

Furthermore, we observe that the ability of neural networks to represent
nonlinear functions enables termination proofs for programs that go beyond
what the state of the art can handle.  Programs that use disjunctive or
nonlinear loop guards as well as programs that require piecewise linear
ranking functions are proven terminating just as easily by our new
technique.

While we perform our experiments with Java programs, our training procedure
is agnostic with respect to the programming language and requires no
information about the program other than execution traces.  It applies
without modifications to software that constructs data structures as long as
a procedure for verifying the candidate ranking functions is available.  The
complexity of formal reasoning about the program is entirely delegated to
the verification procedure, which only has to solve the task of checking the
validity of a given ranking function.

\section{Illustrative Example}\label{sec:example}

We prove that programs terminate by showing that
their loops admit ranking functions. 
Constructing ranking functions for loops that involve
disjunctive loop guards or involve nonlinear constraints
is hard for existing technologies.
As an exemplar, consider the loop in Fig.~\ref{fig:disj}.
This loop terminates for every arbitrary initialisation of the variables
{\sf x}, {\sf y}, and {\sf z} and only involves linear constraints.
Yet, a linear ranking argument is insufficient to prove that this loop terminates.
Naively one might believe that $-${\sf x} is a ranking argument
because it decreases with every loop iteration;
however, it is not bounded from below by a given constant coefficient.
Notably, the same holds for expression ${\sf y} + {\sf z} - {\sf x}$, which
can be always assigned to a value that is smaller than any given constant
coefficient with an adversary initialisation of {\sf y} and {\sf z},
if we assume that {\sf x}, {\sf y}, and {\sf z} can take any unbounded integer.
In fact, under this assumption, no linear combination of {\sf x}, {\sf y}, and {\sf z} is a
valid ranking function for this loop, which requires a nonlinear ranking function.

\begin{figure}
  \centering
\begin{tikzpicture}
  % input layer
  \node (x1) [draw, circle, inner sep=2pt, minimum size=0.4cm]
  at (0,2.25) {};
  \node (x2) [draw, circle, inner sep=2pt, minimum size=0.4cm]
  at (0,1.5) {};
  \node (x3) [draw, circle, inner sep=2pt, minimum size=0.4cm]
  at (0,.75) {};
  
  \def\labeldist{.2cm}
  \node (l1) [left=\labeldist of x1, anchor=east] {\tt x};
  \node (l2) [left=\labeldist of x2, anchor=east] {\tt y};
  \node (l3) [left=\labeldist of x3, anchor=east] {\tt z};
  
  \draw (l1) edge[->] (x1);
  \draw (l2) edge[->] (x2);
  \draw (l3) edge[->] (x3);

  %hidden layer
  \node (y1) [draw, circle, inner sep=2pt, minimum size=0.4cm] at (2, 1.875) {};
  \node (y2) [draw, circle, inner sep=2pt, minimum size=0.4cm] at (2, 1.125) {};
  \node () [above=.5mm of y1] {\small ReLU};
  \node () [below=.5mm of y2] {\small ReLU};
  
  \draw (x1) edge[->] node[above] {$-1$} (y1);
  \draw (x1) edge[->] node[below,pos=.6] {$-1$} (y2);
  \draw (x2) edge[->] node[below,pos=.2] {$1$} (y1);
  \draw (x3) edge[->] node[below] {$1$} (y2);
  
  % output layer
  \node (z) [draw, circle, inner sep=2pt, minimum size=0.4cm] at (3.5,1.5) {};

  \draw (y1) edge[->] node[above, pos=.4] {$1$} (z);
  \draw (y2) edge[->] node[above, pos=.3] {$1$} (z);

  \node (o) [right=\labeldist of z, anchor=west, inner sep=0] {};
  \draw (z) edge[->] (o);
\end{tikzpicture}
\caption{Neural ranking function for the program in Fig.~\ref{fig:disj}.}
\label{fig:exnn}
\end{figure}
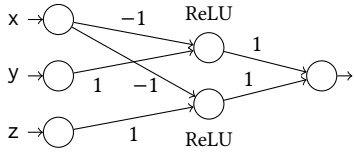

A valid ranking function for the program in Fig.~\ref{fig:disj} is
\begin{equation}
  f({\sf x,y,z}) = \max\{ {\sf y - x}, 0\} + \max \{{\sf z - x}, 0\}.\label{eq:exnn}
\end{equation}
This function not only decreases in every iteration, but is also
non-negative for every valuation of {\sf x}, {\sf y}, and {\sf z} and 
it is bounded from below by zero.  This function corresponds to the
simple neural network with ReLU activation functions in Fig.~\ref{fig:exnn}, which is
effectively learned and verified by our method.

We argue that (1) neural networks are a powerful model to represent ranking
functions of non-trivial programs.  Loops that include disjunctions in their
loop guards are not rare.  Similar behaviour can be induced by conditional
control flow, early breaks or by throwing exceptions.  Suffice it to say
that the following loop is semantically equivalent to the example in Fig.~\ref{fig:disj}:
\begin{minted}{Java}
while (true) {
  if (x >= y && x >= z)
    break;
  x++;
}
\end{minted}

Moreover, we also argue that (2) decoupling the process of guessing
the ranking function from that of checking it enables us to
effectively discover termination arguments for programs
that involve nonlinear constraints. As it turns out,
as of today, neither AProVE nor Ultimate can determine
(within a time budget of 60\,s)
that the following loop terminates:   
\begin{minted}{Java}
while (x*x*x < y) {
  x++;
}
\end{minted}
By contrast, our method learns and verifies
the ranking function $\max\{ {\sf y} - {\sf x}, 0 \}$ in less than a second.

\begin{figure}
  \centering
  \begin{tabular}{cc}
    \begin{minipage}[b]{0.4\columnwidth}
    \begin{minted}{Java}
while (i < k) {
  j = 0;
  while (j < i) {
    j++;
  }
  i++;
}
    \end{minted}
    \end{minipage}
    &
    \begin{tikzpicture}[font=\scriptsize \tt, inner sep=0, minimum size=10pt]
      \node[draw,circle] (l1) {L1};
      \node[draw,circle,below=.2cm of l1] (l2) {L2};
      \node[draw,circle,below=.4cm of l2] (l3) {L3};
      \node[draw,circle,below=.2cm of l3] (l4) {L4};
      \node[draw,circle,below=.2cm of l4] (l5) {L5};
      \node[draw,circle,left=1cm of l2] (l6) {L6};
      \path (l1) edge[->]
      %% node[right] {
      %%   \begin{tabular}{l}iload\_0\\iload\_2\end{tabular}}
      (l2);
      \draw (l2) edge[->] (l6);
      \path (l2) edge[->] (l3);
      \path (l3) edge[->] node[right]
            {} (l4);
      \path (l4) edge[->] (l5);
      \path[draw] (l4) -- 
           ($(l4)-(1.4,0)$)
      -- ($(l3)-(1.4,0)$) edge[->] (l3);
      \path[draw] (l5) -- %% node[below,xshift=-.2cm]
           %% {\begin{tabular}{l}iinc 0,1\\goto L1\end{tabular}}
           ($(l5)+(1.7,0)$)
           -- ($(l1.north)+(1.7,.2cm)$) -- ($(l1.north)+(0,.2cm)$) edge[->] (l1);

      %% \node[right of=l2] {\begin{tabular}{l}if\_icmpge L6\\iconst\_0\\istore\_1\end{tabular}};
      %% \node[right of=l4] {\begin{tabular}{l}if\_icmpge L5\end{tabular}};
      %% \node[right of=l3,anchor=north west] {\begin{tabular}{l}iload\_1\\iload\_0\end{tabular}};

      \node[right=.1cm of l1,anchor=west] (i1) {iload\_0};
      \node[right=.1cm of l2,anchor=west] (i3) {if\_icmpge L6};     
      \node[anchor=west] (i2) at ($(i1.west)!.5!(i3.west)$){iload\_2};
      \node[right=.1cm of l3,anchor=west] (i6) {iload\_1};
      \node[anchor=west] at ($(i3.west)!.33!(i6.west)$){iconst\_0};
      \node[anchor=west] at ($(i3.west)!.66!(i6.west)$){istore\_1};
      \node[right=.1cm of l4,anchor=west] (i8) {if\_icmpge L5};
      \node[anchor=west] at ($(i6.west)!.5!(i8.west)$){iload\_0};
      \node[left=.1cm of l4, anchor=north east] (i9) {iinc 1,1};
      \node[below=.1cm of i9.west, anchor=north west] {goto L3};
      \node[right=.1cm of l5, anchor=north west] (i11) {iinc 0,1};
      \node[below=.1cm of i11.west, anchor=north west] {goto L1}; 
    \end{tikzpicture}\\
    (a) & (b)
  \end{tabular}
  \caption{A Java program and the respective CFG.}
  \label{fig:cfg}
\end{figure}
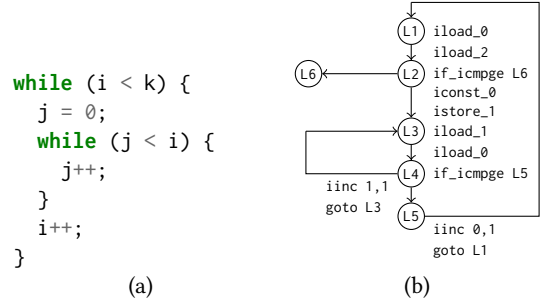 
 
 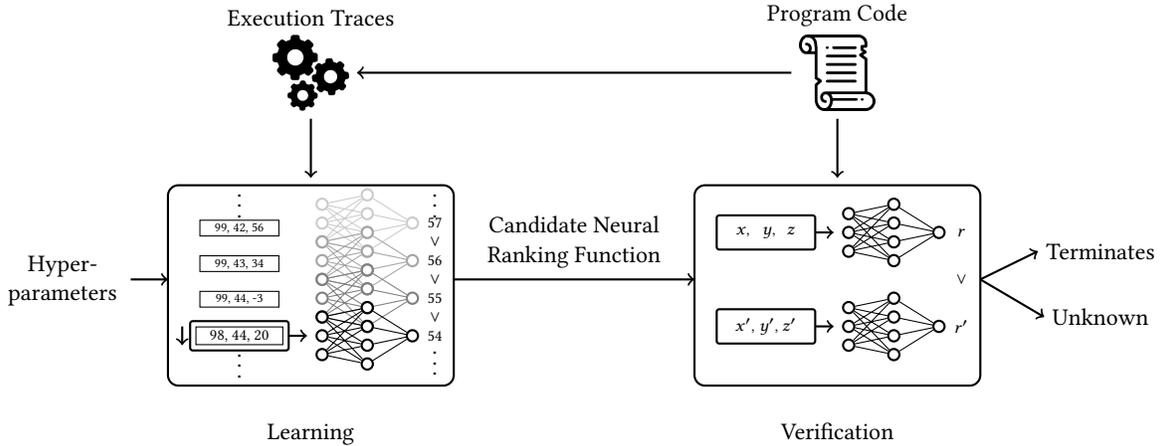
\begin{figure*}[t]
  \centering
  \begin{tikzpicture}[scale=1]

    %neural nets
    \def\nnsep{1.5pt}
    \def\nnyd{.25}
    \def\nnxd{.6}
    \def\nnx{0.15}

    \def\nncolor{black!20}
    \def\nny{6*\nnyd}
    \node[draw, circle, thick, inner sep=\nnsep,\nncolor] (i1) at (\nnx,\nny+\nnyd) {};
    \node[draw, circle, thick, inner sep=\nnsep,\nncolor] (i2) at (\nnx,\nny) {};
    \node[draw, circle, thick, inner sep=\nnsep,\nncolor] (i3) at (\nnx,\nny-\nnyd) {};
    \node[draw, circle, thick, inner sep=\nnsep,\nncolor] (h1) at (\nnx+\nnxd,\nny+1.5*\nnyd) {};
    \node[draw, circle, thick, inner sep=\nnsep,\nncolor] (h2) at (\nnx+\nnxd,\nny+.5*\nnyd) {};
    \node[draw, circle, thick, inner sep=\nnsep,\nncolor] (h3) at (\nnx+\nnxd,\nny-.5*\nnyd) {};
    \node[draw, circle, thick, inner sep=\nnsep,\nncolor] (h4) at (\nnx+\nnxd,\nny-1.5*\nnyd) {};
    \node[draw, circle, thick, inner sep=\nnsep,\nncolor] (o1) at (\nnx+2*\nnxd,\nny) {};
    \draw[\nncolor] (i1) -- (h1) -- (i2) -- (h2) -- (i3) -- (h3) -- (o1) -- (h1)
    -- (i3) -- (h4) -- (o1) -- (h2) -- (i1) -- (h3) -- (i2) -- (h4) -- (i1);
    
    \def\nncolor{black!35}
    \def\nny{4*\nnyd}
    \node[draw, circle, thick, inner sep=\nnsep,\nncolor] (i1) at (\nnx,\nny+\nnyd) {};
    \node[draw, circle, thick, inner sep=\nnsep,\nncolor] (i2) at (\nnx,\nny) {};
    \node[draw, circle, thick, inner sep=\nnsep,\nncolor] (i3) at (\nnx,\nny-\nnyd) {};
    \node[draw, circle, thick, inner sep=\nnsep,\nncolor] (h1) at (\nnx+\nnxd,\nny+1.5*\nnyd) {};
    \node[draw, circle, thick, inner sep=\nnsep,\nncolor] (h2) at (\nnx+\nnxd,\nny+.5*\nnyd) {};
    \node[draw, circle, thick, inner sep=\nnsep,\nncolor] (h3) at (\nnx+\nnxd,\nny-.5*\nnyd) {};
    \node[draw, circle, thick, inner sep=\nnsep,\nncolor] (h4) at (\nnx+\nnxd,\nny-1.5*\nnyd) {};
    \node[draw, circle, thick, inner sep=\nnsep,\nncolor] (o1) at (\nnx+2*\nnxd,\nny) {};
    \draw[\nncolor] (i1) -- (h1) -- (i2) -- (h2) -- (i3) -- (h3) -- (o1) -- (h1)
    -- (i3) -- (h4) -- (o1) -- (h2) -- (i1) -- (h3) -- (i2) -- (h4) -- (i1);
    
    \def\nncolor{black!50}
    \def\nny{2*\nnyd}
    \node[draw, circle, thick, inner sep=\nnsep,\nncolor] (i1) at (\nnx,\nny+\nnyd) {};
    \node[draw, circle, thick, inner sep=\nnsep,\nncolor] (i2) at (\nnx,\nny) {};
    \node[draw, circle, thick, inner sep=\nnsep,\nncolor] (i3) at (\nnx,\nny-\nnyd) {};
    \node[draw, circle, thick, inner sep=\nnsep,\nncolor] (h1) at (\nnx+\nnxd,\nny+1.5*\nnyd) {};
    \node[draw, circle, thick, inner sep=\nnsep,\nncolor] (h2) at (\nnx+\nnxd,\nny+.5*\nnyd) {};
    \node[draw, circle, thick, inner sep=\nnsep,\nncolor] (h3) at (\nnx+\nnxd,\nny-.5*\nnyd) {};
    \node[draw, circle, thick, inner sep=\nnsep,\nncolor] (h4) at (\nnx+\nnxd,\nny-1.5*\nnyd) {};
    \node[draw, circle, thick, inner sep=\nnsep,\nncolor] (o1) at (\nnx+2*\nnxd,\nny) {};
    \draw[\nncolor] (i1) -- (h1) -- (i2) -- (h2) -- (i3) -- (h3) -- (o1) -- (h1)
    -- (i3) -- (h4) -- (o1) -- (h2) -- (i1) -- (h3) -- (i2) -- (h4) -- (i1);

    \def\nncolor{black}
    \def\nny{0}
    \node[draw, circle, thick, inner sep=\nnsep,\nncolor] (i1) at (\nnx,\nny+\nnyd) {};
    \node[draw, circle, thick, inner sep=\nnsep,\nncolor] (i2) at (\nnx,\nny) {};
    \node[draw, circle, thick, inner sep=\nnsep,\nncolor] (i3) at (\nnx,\nny-\nnyd) {};
    \node[draw, circle, thick, inner sep=\nnsep,\nncolor] (h1) at (\nnx+\nnxd,\nny+1.5*\nnyd) {};
    \node[draw, circle, thick, inner sep=\nnsep,\nncolor] (h2) at (\nnx+\nnxd,\nny+.5*\nnyd) {};
    \node[draw, circle, thick, inner sep=\nnsep,\nncolor] (h3) at (\nnx+\nnxd,\nny-.5*\nnyd) {};
    \node[draw, circle, thick, inner sep=\nnsep,\nncolor] (h4) at (\nnx+\nnxd,\nny-1.5*\nnyd) {};
    \node[draw, circle, thick, inner sep=\nnsep,\nncolor] (o1) at (\nnx+2*\nnxd,\nny) {};
    \draw[\nncolor] (i1) -- (h1) -- (i2) -- (h2) -- (i3) -- (h3) -- (o1) -- (h1)
    -- (i3) -- (h4) -- (o1) -- (h2) -- (i1) -- (h3) -- (i2) -- (h4) -- (i1);

    %% sampled states
    \def\trsep{1.1pt}
    \def\tryd{.07}
    \def\trxd{1.1}
    \def\trw{1.05cm}
    \node[draw, rectangle, inner sep=\trsep, minimum width=\trw*1.1] (ta)
    at (\nnx-\trxd, 0) {\scriptsize 98, 44, 20};

    \def\winm{2pt}
    \draw[rounded corners=1pt,thick] ($(ta.north west)+(-\winm,\winm)$) --
    ($(ta.north east)+(\winm,\winm)$) --
    ($(ta.south east)+(\winm,-\winm)$) --
    ($(ta.south west)+(-\winm,-\winm)$) -- cycle;
    
    \draw[thick] ($(ta.east) + (\winm,0)$) edge[->] ($(\nnx,0)-(.2,0)$);

    \draw ($(ta.north west)-(5pt,0)$) edge[thick,->] ($(ta.south west)-(5pt,0)$);
    
    \node[draw, rectangle, inner sep=\trsep, minimum width=\trw,
    above=\nnyd of ta] (ta1) {\tiny 99, 44, -3};
    \node[draw, rectangle, inner sep=\trsep, minimum width=\trw,
      above=\nnyd of ta1] (ta2) {\tiny 99, 43, 34};
    \node[draw, rectangle, inner sep=\trsep, minimum width=\trw,
      above=\nnyd of ta2] (ta3) {\tiny 99, 42, 56};
    \node[above=.02 of ta3, inner sep=0] (t0) {\vdots};
    
    %% \node[draw, rectangle, inner sep=\trsep, minimum width=\trw,
    %% below=\nnyd of ta] (tb1) {\tiny 98, 45, -12};
    %% \node[draw, rectangle, inner sep=\trsep, minimum width=\trw,
    %%   below=\nnyd of tb1] (tb2) {\tiny 97, 45, -46};
    \node[below=-.1 of ta, inner sep=0] (tlast) {\vdots};

    %outputs of nn
    \node at (\nnx+2.5*\nnxd,\nny+6*\nnyd) {\scriptsize 57};
    \node at (\nnx+2.5*\nnxd,\nny+4*\nnyd) {\scriptsize 56};
    \node at (\nnx+2.5*\nnxd,\nny+2*\nnyd) {\scriptsize 55};
    \node at (\nnx+2.5*\nnxd,\nny) {\scriptsize 54};
    \node[rotate=90] at (\nnx+2.5*\nnxd,\nny+\nnyd) {\scriptsize $<$};
    \node[rotate=90] at (\nnx+2.5*\nnxd,\nny+3*\nnyd) {\scriptsize $<$};
    \node[rotate=90] at (\nnx+2.5*\nnxd,\nny+5*\nnyd) {\scriptsize $<$};
    \draw let \p1=(tlast) in node at (\nnx+2.5*\nnxd,\y1+1) {$\vdots$};
    \draw let \p1=(t0) in node at (\nnx+2.5*\nnxd,\y1) {$\vdots$};

    %border about learning
    \draw[thick,rounded corners] let \p1=($(t0.north)$), \p2=($(tlast.south)$) in
    (-1.9, \y2-.1cm) -- (1.9, \y2-.1cm) -- (1.9, \y1-.1cm) --
    (-1.9, \y1-.1cm) -- cycle;

    %border about verification
    \def\verboxsh{7}
    \draw[thick,rounded corners] let \p1=($(t0.north)$), \p2=($(tlast.south)$) in
    (\verboxsh-1.9, \y2-.1cm) -- (\verboxsh+1.9, \y2-.1cm) -- (\verboxsh+1.9, \y1-.1cm) --
    (\verboxsh-1.9, \y1-.1cm) -- cycle;

    %symbolic networks
    \def\nncolor{black}
    \def\nny{0.5*\nnyd}
    \node[draw, circle, thick, inner sep=\nnsep,\nncolor] (i1) at (\nnx+\verboxsh,\nny+\nnyd) {};
    \node[draw, circle, thick, inner sep=\nnsep,\nncolor] (i2) at (\nnx+\verboxsh,\nny) {};
    \node[draw, circle, thick, inner sep=\nnsep,\nncolor] (i3) at (\nnx+\verboxsh,\nny-\nnyd) {};
    \node[draw, circle, thick, inner sep=\nnsep,\nncolor] (h1) at (\nnx+\verboxsh+\nnxd,\nny+1.5*\nnyd) {};
    \node[draw, circle, thick, inner sep=\nnsep,\nncolor] (h2) at (\nnx+\verboxsh+\nnxd,\nny+.5*\nnyd) {};
    \node[draw, circle, thick, inner sep=\nnsep,\nncolor] (h3) at (\nnx+\verboxsh+\nnxd,\nny-.5*\nnyd) {};
    \node[draw, circle, thick, inner sep=\nnsep,\nncolor] (h4) at (\nnx+\verboxsh+\nnxd,\nny-1.5*\nnyd) {};
    \node[draw, circle, thick, inner sep=\nnsep,\nncolor] (o1) at (\nnx+\verboxsh+2*\nnxd,\nny) {};
    \draw[\nncolor] (i1) -- (h1) -- (i2) -- (h2) -- (i3) -- (h3) -- (o1) -- (h1)
    -- (i3) -- (h4) -- (o1) -- (h2) -- (i1) -- (h3) -- (i2) -- (h4) -- (i1);

    \node[inner sep=\trsep, minimum width=\trw*1.1] (ta)
    at (\nnx-\trxd+\verboxsh, \nny) {\scriptsize $x'$, $y'$, $z'$};
    \draw[rounded corners=1pt,thick] ($(ta.north west)+(-\winm,\winm)$) --
    ($(ta.north east)+(\winm,\winm)$) --
    ($(ta.south east)+(\winm,-\winm)$) --
    ($(ta.south west)+(-\winm,-\winm)$) -- cycle;
    \draw[thick] ($(ta.east) + (\winm,0)$) edge[->] ($(\nnx+\verboxsh,\nny)-(.2,0)$);

    \node at (\nnx+2.5*\nnxd+\verboxsh,\nny) {\scriptsize $r'$};
    \node[rotate=90] at (\nnx+2.5*\nnxd+\verboxsh,\nny+2.5*\nnyd) {\scriptsize $<$};

    \def\nncolor{black}
    \def\nny{5.5*\nnyd}
    \node[draw, circle, thick, inner sep=\nnsep,\nncolor] (i1) at (\nnx+\verboxsh,\nny+\nnyd) {};
    \node[draw, circle, thick, inner sep=\nnsep,\nncolor] (i2) at (\nnx+\verboxsh,\nny) {};
    \node[draw, circle, thick, inner sep=\nnsep,\nncolor] (i3) at (\nnx+\verboxsh,\nny-\nnyd) {};
    \node[draw, circle, thick, inner sep=\nnsep,\nncolor] (h1) at (\nnx+\verboxsh+\nnxd,\nny+1.5*\nnyd) {};
    \node[draw, circle, thick, inner sep=\nnsep,\nncolor] (h2) at (\nnx+\verboxsh+\nnxd,\nny+.5*\nnyd) {};
    \node[draw, circle, thick, inner sep=\nnsep,\nncolor] (h3) at (\nnx+\verboxsh+\nnxd,\nny-.5*\nnyd) {};
    \node[draw, circle, thick, inner sep=\nnsep,\nncolor] (h4) at (\nnx+\verboxsh+\nnxd,\nny-1.5*\nnyd) {};
    \node[draw, circle, thick, inner sep=\nnsep,\nncolor] (o1) at (\nnx+\verboxsh+2*\nnxd,\nny) {};
    \draw[\nncolor] (i1) -- (h1) -- (i2) -- (h2) -- (i3) -- (h3) -- (o1) -- (h1)
    -- (i3) -- (h4) -- (o1) -- (h2) -- (i1) -- (h3) -- (i2) -- (h4) -- (i1);

    \node[inner sep=\trsep+1, minimum width=\trw*1.1] (ta)
    at (\nnx-\trxd+\verboxsh, \nny) {\scriptsize $x$,\;\;\,$y$,\;\;\,$z$};
    \draw[rounded corners=1pt,thick] ($(ta.north west)+(-\winm,\winm)$) --
    ($(ta.north east)+(\winm,\winm)$) --
    ($(ta.south east)+(\winm,-\winm)$) --
    ($(ta.south west)+(-\winm,-\winm)$) -- cycle;
    \draw[thick] ($(ta.east) + (\winm,0)$) edge[->] ($(\nnx+\verboxsh,\nny)-(.2,0)$);

    \node at (\nnx+2.5*\nnxd+\verboxsh,\nny) {\scriptsize $r$};

    %program
    \node (scroll) at (\verboxsh, 3.5){\includegraphics[width=1cm]{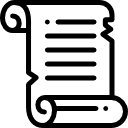}};
    \node [above=-.1cm of scroll] { Program Code };

    %tracer
    \node (run) at (0, 3.5){\includegraphics[width=1cm]{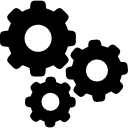}};
    \node [above=-.1cm of run] (tracing) {Execution Traces};
    
    \draw[thick] let \p1=($(t0.north)$) in (run.south) edge[->] (0,\y1);
    \draw[thick] let \p1=($(t0.north)$) in (scroll.south) edge[->] (\verboxsh,\y1);
    \draw[thick] (scroll.west) edge[->] (run.east);
    \draw[thick] (1.9,3*\nnyd) edge[->]
    node[above] {\begin{tabular}{c}
        Candidate Neural\\Ranking Function\end{tabular}}
    (\verboxsh-1.9,3*\nnyd); 

    %output
    \node (yes) at (\verboxsh+3.5, 4.5*\nnyd) { Terminates}; \node (no) at
    (\verboxsh+3.5, 1*\nnyd) { Unknown}; \draw[thick] (\verboxsh+1.9, 3*\nnyd)
    edge[->] (yes.west); \draw[thick] (\verboxsh+1.9, 3*\nnyd) edge[->]
    (no.west);

    % input
    \node[inner sep=0] (input) at (-3.3, 3*\nnyd) { \begin{tabular}{c}
        Hyper-\\parameters\end{tabular}}; \draw[thick] (input) edge[->] (-1.9,
    3*\nnyd);

    \def\capd{1.1} \draw node[anchor=north, inner sep=0] at (0,-\capd)
    {\begin{tabular}{c}Learning\end{tabular}}; \draw node[anchor=north, inner
    sep=0] at (\verboxsh, -\capd) {\begin{tabular}{c}Verification\end{tabular}};
  \end{tikzpicture}
  \caption{Schema of our framework.}
  \label{fig:arch}
\end{figure*}

\section{Background}\label{sec:ranking-nn}\label{sec:learning-ranking-functions} \label{sec:term-analysis-ml}\label{sec:prelims}

\paragraph{Programs and Transition Systems}

A computer program is a list of instructions that, together with the machine
interpreting them, defines a state transition system over the state space of
the machine.  A state transition system is a pair $P = (S,T)$ where $S$ is a
countable (possibly infinite) set of states and $T \subset S \times S$ is a
transition relation. A state contains all information that is necessary to
determine its successor state(s). For example, this can include the value
of variables that are explicitly declared in the source code, or
that exist as registers in the interpreter (e.g., the program counter).
The transition relation determines the successors of a state.
A state without any successors is a terminating state.
A state may have multiple successors because of operations that
are external to the program (e.g., non-deterministic input assignments)
or are underspecified and may therefore have multiple outcomes. 
A~run of $P$ is any sequence of states
%
%% \begin{equation}
  $s^{(0)}, s^{(1)}, s^{(2)}, \dots$ %\label{eq:run}
%% \end{equation}
%
such that $(s^{(i)},s^{(i+1)}) \in T$ for all $i \geq 0$.
We say that a program terminates if all its runs are finite.

\paragraph{Control Flow Graphs and Loop Headers} A control flow graph (CFG) for program $P$
is a finite directed graph $G = (L, E)$ where $L$ is a finite set
of control locations and $E \subseteq L \times L$ is a set of control edges.
For Java programs, a CFG can be obtained for its bytecode,
as illustrated in Fig.~\ref{fig:cfg}.
Control locations correspond to source and target addresses of jump instructions
or entry or exit points of a procedure. Control edges indicate
whether there exists a sequence of instructions or a jump that lead from
the respective source to the respective destination location.
A state is on a control location if the next instruction to be
executed is on a control location; in the Java Virtual Machine,
this is determined by the program counter. Notably, we have that 
for every finite run $s^{(0)}, \dots, s^{(k)}$ such
that $s^{(0)}$ and $s^{(k)}$ are respectively on control locations $l$ and $l'$,
the control graph must admit a path from $l$ to $l'$.
Then, let $L' \subseteq L$ be any subset of control locations.
We define $S_{L'} \subseteq S$ as the set of states on any location in $L'$
and $T_{L'} \subseteq S_{L'} \times S_{L'}$ as the
the maximal relation of states on control locations in $L'$
such that $(s^{(0)}, s^{(k)}) \in T_{L'}$ only if there exist
a finite run $s^{(0)}, s^{(1)}, \dots, s^{(k-1)}, s^{(k)}$ 
that does not encounter any control location in $L'$ in between, i.e., 
$s^{(1)}, \dots, s^{(k-1)} \not\in S_{L'}$. 
Special control locations are loop headers, that is, the dominators (entry locations) of the
strongly connected components in the graph~\cite{cfg}; for instance, the
loop headers for Fig.~\ref{fig:cfg}b are L1 and L3.
They are important in termination analysis because every run along a
loop (i.e., {\sf for} and {\sf while} statements) 
necessarily enters and iterates over at least one of them.
We denote the set of loop headers as $H \subseteq L$.

\paragraph{Ranking Functions} 

To determine whether a program terminates, our method attempts to find a
ranking function for it. A function $f \colon S_H \to R$
is a ranking function for the program $P$ if
\begin{equation}
  f(s') \prec f(s)\quad \text{for all } (s,s') \in T_H
\end{equation}
and the relation $(R,\prec)$ is well founded~\cite{Floyd67}.  The existence
of a ranking function proves that the program terminates, and every program
that does not terminate necessarily lacks a ranking function.  A standard
way of giving a ranking function is to identify a function that maps states
at loop headers to sequences of numbers that (i)~decrease by a discrete
amount and (ii)~are bounded from below.  Another way is to define tuples of
functions that decrease lexicographically at loop headers, which is
particularly useful for nested loops.

\section{Overview of the Method}\label{sec:overview}
We propose a framework (\prettyref{fig:arch}) for termination analysis using
neural networks as ranking functions. These neural ranking functions
are first trained over execution traces of a program and
subsequently verified in combination with the program code.
Thus, the three parameters of \emph{neural termination analysis} are
\begin{enumerate}
\item the tracing/sampling strategy,
\item the neural network architecture, and
\item the verification procedure.
\end{enumerate}
These steps are independent from each other. Hence, changing the learning setup
(e.g. considering different models, other learning frameworks, etc.) only requires a change in the second
step. Similarly, considering different input languages or alternative
verification procedures
only requires a change in
the verification and (potentially) tracing procedure.

As illustrated in \prettyref{fig:arch}, in the first step we collect
execution traces for a program $P$ that we want to show terminating.  These
execution traces are used subsequently as training data to train the neural
ranking functions.  These traces are obtained by first generating test input
data for $P$ and then tracing the execution of $P$ with the test input data.
Since the training phase exclusively works with these execution traces it is
important that they adequately represent the behaviour of the
program~$P$. More details can be found in \prettyref{appendix:tracing}.

In the second step, we use the execution traces to train neural networks to
become neural ranking functions.  We discuss suitable choices for neural
network architectures in \prettyref{sec:learning-ranking-functions}.  We
always use neural architectures that guarantee that the neural network's
output is bounded from below.  The training procedure minimises a loss
function that punishes neural networks that do not decrease over the sampled
observations.  As a result, we obtain a neural network that behaves like a
ranking function over the sampled executions traces.

Finally, to assert that a trained neural ranking function generalises to all possible
executions, we pass it to a formal verification procedure.  The formal
verification procedure encodes both the program and the neural network
symbolically (more details about our encoding are shown in \prettyref{appendix:verification}).
Then, it uses SMT solving to formally
decide whether the neural network is a valid ranking function for the program.
Upon an affirmative result, we conclude that the program terminates; upon a negative
result, we return an inconclusive answer, that is, the program may or may
not terminate.  The verification procedure guarantees that our method is
sound. 

The combination of a particular sampling strategy, neural architecture, and
verification procedure is an instance of our approach, which offers a
flexible and extensible termination analysis framework that combines testing
methods, deep learning, and symbolic reasoning.

\section{Neural Ranking Functions}\label{sec:learning-ranking-functions}\label{sec:learning}

Our method collects execution traces for a program and then proceeds in two
phases: first it trains and then it formally verifies.  The first phase thus
takes a set of execution traces as input and returns a neural network as
output.  The set of execution traces forms the dataset which is used to
train a neural network that mimics a ranking function along these traces. 
The neural network is trained by minimising a loss function that ensures
that the neural network decreases by a discrete amount after every pair of
subsequent observations in the traces.  We analyse every loop in the program
and provide a ranking function for each of them.

We provide two strategies for training these candidate neural ranking
function. The first strategy trains a monolithic ranking argument, that is,
a neural ranking function that outputs one value that decreases along the
traces.  The second strategy generalises the first and trains a
lexicographic ranking argument, that is, a neural ranking function that
outputs many values that decrease lexicographically.  Whether to use one or
the other strategy is heuristic; lexicographic arguments are normally
suitable for nested loops.

\paragraph{Observation Functions and Traces}
We train our candidate neural ranking function from states collected
along program runs. 
However, this is made difficult by the fact that the states of a program
are large and complex and contain internal
information that is not directly amenable to deep learning.
The standard approach to address this issue is
to construct an \emph{embedding},
defined by means of an observation function.
An observation function $\omega \colon S_{L'} \to \mathbb{R}^n$ extracts
vectors of numerical values that can be taken as input by a neural network,
from states that are on a specific set of control locations $L' \subseteq L$.
These numerical values can be, for instance,
the values of numerical variables in memory.
With an observation function $\omega$ we convert any run into a trace
$o^{(0)}, o^{(1)}, o^{(2)}, \dots$, which is the sequence of
observations in $\Real^n$ recorded every time a location in $L'$ is encountered.  
In other words, every such trace corresponds to a sequence
$s^{(0)}, s^{(1)}, s^{(2)}, \dots$ of states in $S_{L'}$ such that
$o^{(i)} = \omega(s^{(i)})$ and
$(s^{(i)},s^{(i+1)}) \in T_{L'}$, for all $i \geq 0$. 

\paragraph{General Architecture for Neural Ranking Functions}
We use feed-forward neural networks as models of ranking functions.
Generally, this is a function $f \colon \Theta \times \Real^n \to \Real^m$ with $n$ inputs,
$m$ outputs, and parameterised by $\theta \in \Theta$, defined in terms of interconnected neurons
partitioned into one input, one output, and $k$ intermediate hidden layers.
The intermediate hidden layers are defined as a parameterised function
\begin{equation}
  \sigma(W, b; x) = \relu(Wx+b)
\end{equation}
where the parameters are a matrix of weights $W$ and a vector of biases $b$.
These define an affine transformation of the input, 
while $\relu$ applies a nonlinear transformation $\max\{ \cdot, 0 \}$
element-wise to each of the $h$ neurons in the respective hidden layer, that is
\begin{equation}
  \relu(x_1, \dots, x_h) = (\max\{x_1, 0\}, \dots, \max\{x_h, 0\}).
\end{equation}
The parameters of the network are thus a sequence of weight matrices
and bias vectors $\theta = (W^{(1)}, b^{(1)}, \dots, W^{(k)}, b^{(k)})$, each of
which corresponds to a hidden layer. We impose that the output layer has no bias,
its weights $W^{(k+1)}$ are not trainable---and their coefficients are non-negative.
The neural network thus defines the function  
\begin{equation}
  f(\theta;\,\cdot\,) = W^{(k+1)} \sigma(W^{(k)}, b^{(k)};\,\cdot\,) \circ \dots \circ
    \sigma(W^{(1)}, b^{(1)};\,\cdot\,),
\end{equation}
whose output is in turn guaranteed to be non-negative for every valuation of inputs and parameters.
This ensures that the function output is always bounded from below by zero.
To train this neural network so as it behaves as a raking function it
remains to ensure that it decreases after every loop iteration.

\begin{figure}
  \centering
  \begin{tikzpicture}[scale=0.9,inner sep=2pt, minimum size=0.4cm,node
    distance=.5cm]

    % input layer
    \node (x1) [draw, circle] at (0,.6) {};
    \node (x2) [draw, circle, below of=x1] {};

    \node (x3) [inner sep=0pt,minimum size=0,below of=x2] {\vdots};
    \node (x4) [draw, circle, below of=x3,yshift=-2mm] {};
    
    \def\labeldist{.4cm} \node (l1) [left=\labeldist of x1, anchor=east]
    {$x_1$}; \node (l2) [left=\labeldist of x2, anchor=east] {$x_2$}; \node (l4)
    [left=\labeldist of x4, anchor=east] {$x_n$};
    
    \draw (l1) edge[->] (x1); \draw (l2) edge[->] (x2); \draw (l4) edge[->]
    (x4);
    
    % 1st hidden layer
    \node (y1) [draw, circle] at (1.5,.85) {};
    % \node [above=.01cm of y1] {ReLU};
    \node (y2) [draw, circle, below of=y1] {};
    % \node [above=.01cm of y2] {ReLU};
    \node (y3) [draw, circle, below of=y2] {};
    \node (y4) [inner sep=0pt,minimum size=0,below of=y3,yshift=.1cm] {\vdots};
    \node (y5) [draw, circle, below of=y4,yshift=-.1cm] {};

    \def\bmar{3pt} \draw[rounded corners] let
    \p1=(y1.north),\p2=(y1.east),\p3=(y5.south),\p4=(y5.west) in
    (\x2+\bmar,\y1+\bmar) -- (\x2+\bmar,\y3-\bmar) -- (\x4-\bmar,\y3-\bmar) --
    (\x4-\bmar,\y1+\bmar) -- cycle;

    % 2nd hidden layer
    %\node (y21) [draw, circle] at (3,.85) {};
    % \node [above=.01cm of y1] {ReLU};
    %\node (y22) [draw, circle, below of=y21] {};
    % \node [above=.01cm of y2] {ReLU};
    %\node (y23) [draw, circle, below of=y22] {};
    %\node (y24) [inner sep=0pt,minimum size=0,below of=y23,yshift=.1cm] {\vdots};
    %\node (y25) [draw, circle, below of=y24,yshift=-.1cm] {};
    
    %\def\bmar{3pt} \draw[rounded corners] let
    %\p1=(y21.north),\p2=(y21.east),\p3=(y25.south),\p4=(y25.west) in
    %(\x2+\bmar,\y1+\bmar) -- (\x2+\bmar,\y3-\bmar) -- (\x4-\bmar,\y3-\bmar) --
    %(\x4-\bmar,\y1+\bmar) -- cycle;

    % 3rd hidden layer
    \draw let \p1=(y3) in node at (2.4,\y1) {$\dots$};

    % 4th hidden layer
    \node (z1) [draw, circle] at (3.2,.85) {};
    % \node [above=.01cm of y1] {ReLU};
    \node (z2) [draw, circle, below of=z1] {};
    % \node [above=.01cm of y2] {ReLU};
    \node (z3) [draw, circle, below of=z2] {};
    \node (z4) [inner sep=0pt,minimum size=0,below of=z3,yshift=.1cm] {\vdots};
    \node (z5) [draw, circle, below of=z4,yshift=-.1cm] {};
    
    \def\bmar{3pt} \draw[rounded corners] let
    \p1=(z1.north),\p2=(z1.east),\p3=(z5.south),\p4=(z5.west) in
    (\x2+\bmar,\y1+\bmar) -- (\x2+\bmar,\y3-\bmar) -- (\x4-\bmar,\y3-\bmar) --
    (\x4-\bmar,\y1+\bmar) -- cycle;

    %output node
    \node (o) [draw, circle, right=1.3 of z3]  {};

    \foreach \x in {1,2,4}{
        \foreach \y in {1,2,3,5}{
            \draw (x\x) -- (y\y);
        }
      }

    %\foreach \x in {1,2,3,5}{
    %    \foreach \y in {1,2,3,5}{
    %        \draw (y\x) -- (y2\y);
    %    }
    %}
    
    \foreach \x in {1,2,3,5}{
      \draw (z\x) -- node[above,pos=.3]{\small 1} (o);
    }

    \node (ol) [right=\labeldist of o, anchor=west] {$f(\theta;x)$};
    \draw (o) edge[->] (ol);

    \node [above of=y1] {\small $\relu$};
    %\node [above of=y21] {\small $\relu$};
    \node [above of=z1] {\small $\relu$};
    
    \def\brdist{3mm}
    \draw[decorate,thick,decoration={brace,amplitude=6pt,mirror,raise=0pt}] let
    \p1=(x1.east),\p2=(z5.west),\p3=(z5.south) in (\x1,\y3-\brdist) --
    (\x2-2*\bmar,\y3-\brdist) node[midway,yshift=-4mm]{\small
      Trainable parameters $\theta$};
  \end{tikzpicture}
  \caption{Architecture for monolithic NRFs.}
  \label{fig:monn}
\end{figure}
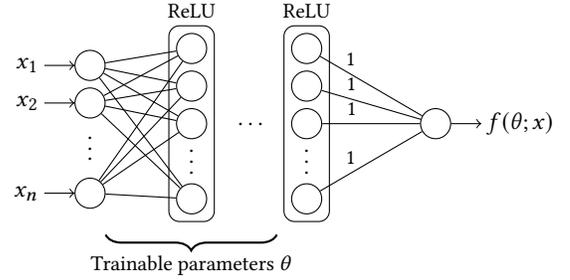
\paragraph{Monolithic Ranking Loss}
A monolithic neural ranking function is the special case where $m=1$;
one example is depicted in Fig.~\ref{fig:monn}.
Our goal is to train it in such a way its output decreases by a discrete amount $\delta > 0$
along a set of sample traces of observations collected every time a loop header is encountered.
For this purpose, we define an embedding using an 
observation function $\omega \colon S_H \to \Real^n$.
With this embedding, we record multiple traces from the program under analysis and
store a dataset of observation pairs $D \subset \Real^n \times \Real^n$ such
that for every pair $(\bm{o},\bm{o}')$ we have that
$\bm{o}$ is immediately followed by
$\bm{o}'$ in the trace. In other words, $D$ constitutes a sliding window of
size two over the trace.
Note that $D$ possibly contains the pairs of multiple execution traces. 
We train our network so as to decrease between every pair $(\bm{o},\bm{o}') \in D$
of sampled observations:
\begin{equation}
  f(\theta;\bm{o}') \leq f(\theta;\bm{o}) - \delta.
\end{equation}
To this end, we solve the following optimisation problem:  
\begin{equation}
  \argmin_{\theta} \frac{1}{|D|}
  \sum_{(\bm{o},\bm{o}') \in D} \underbrace{\max\{f(\theta;\bm{o}') - f(\theta;\bm{o}) + \delta, 0 \}}_{\mathcal{L}(\bm{o}, \bm{o}', \theta)}\label{eq:monolithicopt}
\end{equation}
Function $\mathcal{L}(\bm{o}, \bm{o}', \theta)$ is the loss of the neural
network over a given pair.  The higher the value of $\mathcal{L}$ the more
the network increases over the pair, whereas for pairs that decrease by at
least $\delta$ the value is always 0 (i.e., negative values do not affect
the sum).  As a result of the optimisation problem, we obtain parameters
that ensure the network decreases along all sampled traces.

\begin{example}
  Consider the program in Fig.~\ref{fig:disj}.
  For this program, we define an embedding $\omega \colon S_H \to \Int^3$ that
  observes the values of {\tt x}, {\tt y}, and {\tt z} every time a run
  hits the entry location of the loop. We reason about the loop
  in isolation and sample random initial values for the variables.
  Two example traces are
  \begin{align*}
    &(5,-3,10), (6,-3,10), (7,-3,10), (8,-3,10), (9,-3,10), (10,-3,10)\\
    \intertext{and}
    &(-2,3,-5), (-1,3,-5), (0,3,-5), (1,3,-5), (2,3,-5),(3,3,-5).
  \end{align*}
  The dataset corresponding to exactly these traces thus contains 10 pairs
  of consecutive observations. We use the monolithic architecture in Fig.~\ref{fig:exnn},
  which has exactly one hidden layer made of two neurons.
  We set $\delta = 1$ and observe that once the parameters in Fig.~\ref{fig:exnn} are
  attained, the loss function~$\mathcal{L}$ measures zero over all pairs, which is its minimum.
  The corresponding function $f$ in Eq.~\ref{eq:exnn} maps both traces to the sequence  
  \begin{equation*}
    5,4,3,2,1,0.
  \end{equation*}
  The SMT solver successfully verifies that the following formula evaluates to true for
  every possible assignment to ${\sf x},{\sf y}, {\sf z},\allowbreak {\sf x}',{\sf y}',{\sf z}'$:
  \begin{multline}
    [({\sf x} < {\sf y} \lor {\sf x} < {\sf z}) \land {\sf x}' = {\sf x} + 1 \land {\sf y}' = {\sf y} \land {\sf z}' = {\sf z}] \implies \\
    f({\sf x},{\sf y},{\sf z}) - \delta \geq f({\sf x}',{\sf y}',{\sf z}') 
  \end{multline}
  Unprimed variables represent an observation before an iteration and primed variables after an iteration.
  The validity of this formula confirms that $f$ decreases by $\delta$ for every possible assignment of the variables and
  is thus a valid ranking function. Note that $f$ is bounded from below by
  construction owing to the use of ReLUs.
\end{example}

\begin{figure}
  \centering
  \begin{tikzpicture}[scale=0.9,inner sep=2pt, minimum size=0.4cm,node
    distance=.5cm]
    % input layer
    \node (x1) [draw, circle] at (0,.6) {}; \node (x2) [draw, circle, below
    of=x1] {}; \node (x3) [inner sep=0pt,minimum size=0,below of=x2] {\vdots};
    \node (x4) [draw, circle, below of=x3,yshift=-2mm] {};
    
    \def\labeldist{.4cm} \node (l1) [left=\labeldist of x1, anchor=east]
    {$x_1$}; \node (l2) [left=\labeldist of x2, anchor=east] {$x_2$}; \node (l4)
    [left=\labeldist of x4, anchor=east] {$x_n$};
    
    \draw (l1) edge[->] (x1); \draw (l2) edge[->] (x2); \draw (l4) edge[->]
    (x4);

    % hidden layer
    \node (y1) [draw, circle] at (2,2) {};
    % \node [above=.01cm of y1] {ReLU};
    \node (y2) [draw, circle, below of=y1] {};
    % \node [above=.01cm of y2] {ReLU};
    \node (y3) [inner sep=0pt,minimum size=0,below of=y2,yshift=.1cm] {\vdots};
    \node (y4) [draw, circle, below of=y3,yshift=-.1cm] {};

    \def\bmar{3pt} \draw[rounded corners] let
    \p1=(y1.north),\p2=(y1.east),\p3=(y4.south),\p4=(y4.west) in
    (\x2+\bmar,\y1+\bmar) -- (\x2+\bmar,\y3-\bmar) -- (\x4-\bmar,\y3-\bmar) --
    (\x4-\bmar,\y1+\bmar) -- cycle;

    \node (y5) [inner sep=0pt,minimum size=0,below of=y4] {\vdots};

    \node (y6) [draw, circle, below of=y5,yshift=-.5cm] {}; \node (y7) [draw,
    circle, below of=y6] {}; \node (y8) [inner sep=0pt,minimum size=0,below
    of=y7,yshift=.1cm] {\vdots}; \node (y9) [draw, circle, below
    of=y8,yshift=-.1cm] {};

    \draw[rounded corners] let
    \p1=(y6.north),\p2=(y6.east),\p3=(y9.south),\p4=(y9.west) in
    (\x2+\bmar,\y1+\bmar) -- (\x2+\bmar,\y3-\bmar) -- (\x4-\bmar,\y3-\bmar) --
    (\x4-\bmar,\y1+\bmar) -- cycle;

    %dots
    \draw let \p1=(y1.north),\p3=(y4.south) in node at (2.75,.5*\y1+.5*\y3) {\dots};
    \draw let \p1=(y6.north),\p3=(y9.south) in node at (2.75,.5*\y1+.5*\y3) {\dots};
    
    %2nd hidden layer
    \node (z1) [draw, circle] at (3.5,2) {};
    % \node [above=.01cm of y1] {ReLU};
    \node (z2) [draw, circle, below of=z1] {};
    % \node [above=.01cm of y2] {ReLU};
    \node (z3) [inner sep=0pt,minimum size=0,below of=z2,yshift=.1cm] {\vdots};
    \node (z4) [draw, circle, below of=z3,yshift=-.1cm] {};

    \def\bmar{3pt} \draw[rounded corners] let
    \p1=(z1.north),\p2=(z1.east),\p3=(z4.south),\p4=(z4.west) in
    (\x2+\bmar,\y1+\bmar) -- (\x2+\bmar,\y3-\bmar) -- (\x4-\bmar,\y3-\bmar) --
    (\x4-\bmar,\y1+\bmar) -- cycle;

    \node (z5) [inner sep=0pt,minimum size=0,below of=z4] {\vdots};

    \node (z6) [draw, circle, below of=z5,yshift=-.5cm] {}; \node (z7) [draw,
    circle, below of=z6] {}; \node (z8) [inner sep=0pt,minimum size=0,below
    of=z7,yshift=.1cm] {\vdots}; \node (z9) [draw, circle, below
    of=z8,yshift=-.1cm] {};

    \draw[rounded corners] let
    \p1=(z6.north),\p2=(z6.east),\p3=(z9.south),\p4=(z9.west) in
    (\x2+\bmar,\y1+\bmar) -- (\x2+\bmar,\y3-\bmar) -- (\x4-\bmar,\y3-\bmar) --
    (\x4-\bmar,\y1+\bmar) -- cycle;

    \node [above of=y1] {\small $\relu$};
    \node [above of=y6] {\small $\relu$};
    \node [above of=z1] {\small $\relu$};
    \node [above of=z6] {\small $\relu$};

    % output layer
    \draw let \p1=(y1.north),\p3=(y4.south) in node (o1) [draw, circle, inner
    sep=2pt, minimum size=0.4cm] at (5.5,.5*\y1+.5*\y3) {};

    \draw let \p1=(y6.north),\p3=(y9.south) in node (o2) [draw, circle, inner
    sep=2pt, minimum size=0.4cm] at (5.5,.5*\y1+.5*\y3) {};

    \node at ($(o1)!.5!(o2)$) {\vdots};
    
    \node (ol1) [right=\labeldist of o1, anchor=west] {$f_1(\theta;x)$};
    \draw (o1) edge[->] (ol1);
    
    \node (ol2) [right=\labeldist of o2, anchor=west] {$f_m(\theta;x)$};
    \draw (o2) edge[->] (ol2);

    \foreach \x in {1,2,4}{
        \foreach \y in {1,2,4}{
            \draw (x\x) -- (y\y);
        }
    }

    \foreach \x in {1,2,4}{
        \foreach \y in {6,7,9}{
            \draw (x\x) -- (y\y);
        }
    }

    \foreach \x in {1,2,4}{
      \draw (z\x) -- node[above, pos=.25] {\small 1} (o1);
    }

    \foreach \x in {6,7,9}{
      \draw (z\x) -- node[above, pos=.25] {\small 1} (o2);
    }
    
    \def\brdist{3mm}
    \draw[decorate,thick,decoration={brace,amplitude=6pt,mirror,raise=0pt}] let
    \p1=(x1.east),\p2=(z1.west),\p3=(z9.south) in (\x1,\y3-\brdist) --
    (\x2-2*\bmar,\y3-\brdist) node[midway,yshift=-4mm]{\small
      Trainable parameters $\theta$};
  \end{tikzpicture}
  \caption{Architecture for lexicographic NRFs.}
  \label{fig:lexnn}
\end{figure}
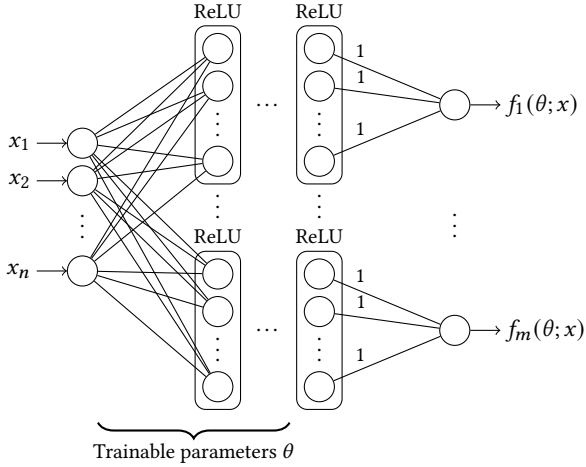

\paragraph{Lexicographic Ranking Loss}

Neural ranking functions with $m \ge 2$ can learn
lexicographic ranking arguments.
Figure~\ref{fig:lexnn} gives an example of an architecture for this purpose. 
Our goal is training the neural network to ensure
that every pair of observations decreases some output neuron by $\delta$,
as long as all other outputs with smaller index do not increase.
Lexicographic arguments are suitable to programs with
multiple loop headers. In this case, we associate to
each loop header an index $i \in \{ 1, \dots, m \}$ for the
output neuron that we expect to decrease every time the header is visited.
For every header we define an embedding 
that records an observation every time that header is visited. 
We thus obtain multiple datasets $D_1, \dots, D_m$ from our sample runs,
one for each header.
We train our network to obtain that, for each pair $(\bm{o}, \bm{o}') \in D_i$ at the $i$-th header,
the output neurons decrease lexicographically as follows:
\begin{align}
  &f_{i}(\theta;\bm{o}') \leq f_{i}(\theta;\bm{o}) - \delta&&\text{and}\label{eq:lex1}\\
  &f_j(\theta;\bm{o}') \leq f_j(\theta;\bm{o})&&\text{for all}~j < i. \label{eq:lex2}
\end{align}
To train the network we solve the following problem:
\begin{equation}
  \argmin_{\theta} \frac{1}{|D_1| + \dots + |D_m|} \sum_{i=1}^m \sum_{(\bm{o},\bm{o}') \in D_i} \mathcal{L}_i(\bm{o},\bm{o}',\theta) \label{eq:lexiopt}  
\end{equation}
The loss of a pair of observations is determined by the dataset it belongs to:
\begin{multline}
  \mathcal{L}_i(\bm{o},\bm{o}',\theta) = \max\{f_{i}(\theta;\bm{o}') - f_{i}(\theta;\bm{o}) + \delta, 0 \} + 
  \\
  \sum_{j = 1}^{i-1} \max\{f_j(\theta;\bm{o}') - f_j(\theta;\bm{o}), 0\}.
\end{multline}
Function $\mathcal{L}_i(\bm{o},\bm{o}',\theta)$ takes its minimal value 0
when both conditions \eqref{eq:lex1} and \eqref{eq:lex2} are satisfied.
If \eqref{eq:lexiopt} also attains value 0, then all samples satisfy these conditions and the
network constitutes a lexicographic neural raking function over the sampled traces.
 
\begin{example}
  Consider the program in Fig.~\ref{fig:cfg}, which has two nested loops.
  The outer loop has its header at location L1 and the inner at location L3.
  We associate L1 with index~1 and L3 with index~2 and
  learn a lexicographic argument accordingly.
  We define two embeddings $\omega_1 \colon S_{\{\sf L1\}} \to \Int^3$ and
  $\omega_2 \colon S_{\{\sf L3\}} \to \Int^3$ that map
  states at control locations {\sf L1}, resp.~{\sf L3},
  to the values of {\tt i}, {\tt j}, and {\tt k}.
  We take, for this example, one sample run with initial values 0,0,3 and obtain
  the following two datasets of pairs induced by $\omega_1$ and $\omega_2$ respectively: 
  \begin{align*}
    &D_1 = \{((0,0,3), (1,0,3)),((1,0,3), (2,1,3)), ((2,1,3), (3,2,3))\}\\
    &D_2 = \{((1,0,3), (1,1,3)), ((2,0,3), (2,1,3)), ((2,1,3), (2,2,3))\} 
  \end{align*}
  We use a neural network as in Fig.~\ref{fig:lexnn} with one hidden layer and one hidden neuron
  in each of the two blocks. The first output must converge to the following function:
  \begin{equation}
    f_1({\sf i, j, k}) = \max \{ {\sf k} - {\sf i}, 0 \}.\label{eq:lex1}
  \end{equation}
  The second output may converge to either of the following functions, which are both valid:
  \begin{align}
    &f_2({\sf i, j, k}) = \max \{ {\sf i} - {\sf j}, 0 \}\\
    &f_2({\sf i, j, k}) = \max \{ {\sf k} - {\sf j}, 0 \}\label{eq:lex2v2}
  \end{align}
  Proving that these functions are a valid termination lexicographic argument relies on
  auxiliary invariants which, as discussed in the appendix, we extract using a
  syntactic heuristic.
  For instance, checking that \eqref{eq:lex2v2} decreases along the inner loop requires
  the auxiliary invariant $i < k$; checking that \eqref{eq:lex1} decreases along the outer loop
  requires an argument that the inner loop leaves {\tt k} and {\tt i} unchanged.
\end{example}

\section{Experiments}\label{sec:case-studies}\label{sec:experim}
\begin{table*}
  \centering
  \begin{tabularx}{\textwidth}{p{0.07\textwidth} p{0.0\textwidth} | X X | X X | X X || X X | X X}
    \cmidrule[\heavyrulewidth]{3-12}
    % \cmidrule{3-5}
    & &\multicolumn{2}{X|}{\texttt{Aprove\_09}(1)} &
    \multicolumn{2}{X|}{\texttt{term-crafted}(2)} &
    \multicolumn{2}{X||}{\texttt{nuTerm\_advantage}(3)} &
    \multicolumn{2}{X|}{combined$\{1,2\}$} &
    \multicolumn{2}{X}{combined$\{1,2,3\}$}\\
    \# & & 38 & & 72& & 14 & &110 & & 124 \\
    \midrule
     \ourTool         & & 34.6 & \phantom{0}91\%   & 49.6 & 68.9\% & 13 & 92.9\% & 84.2 & 76.6\% & 97.2 & 78.4\%\\
    \midrule%
    \texttt{Aprove}   & & 34   & 89.5\% & 64 & 88.9\% & \phantom{0}3 & 21.4\%  & \phantom{0}98 & 89.1\% & 102  &81.5\% \\%
    \midrule%
    \texttt{Ultimate} & & 34   & 89.5\% & 61 & 84.5\% & \phantom{0}1 & \phantom{0}7.8\%  & \phantom{0}95 & 86.4\% & \phantom{0}96 & 77.4\% \\%
    \midrule%
    \texttt{DynamiTe} & & 31   & 81.6\% & 46 & 63.9\% & \phantom{0}7 & \phantom{0}50\% & \phantom{0}77 & \phantom{0}70\% & \phantom{0}84 & 67.7\% \\%
    \bottomrule%
  \end{tabularx}
  \caption{Results of running \ourTool (with 7 neurons), \texttt{AProVE}, \texttt{Ultimate}, and
    \texttt{DynamiTe} on \texttt{term-crafted},
    \texttt{Aprove\_09}, and \texttt{nuTerm-advantage}. In the case of \ourTool
    we report the average results rounded to the first decimal. The last two columns show the union
    of the first two problem sets and all three problem sets, respectively.}
  \label{tab:results}
\end{table*}

We present an experimental evaluation to answer the following research
questions:
\begin{description}
\item[RQ1] Can neural ranking functions be used to formally prove the
termination of programs?
\item[RQ2] Do neural ranking functions advance the state of the art in
termination analysis?
\item[RQ3] How do neural ranking functions scale in terms of the complexity
of the program?
\end{description}
To answer these questions, we developed a prototype implementation of the
methods discussed in the previous sections for proving termination of Java
programs, which we name \ourTool.  Our implementation strictly separates
tracing, learning, and verification as discussed in
\prettyref{sec:overview}.

To obtain program traces we first generate test data for the program in
question using a multivariate normal distribution.  Subsequently, we execute
the program using this test data while maintaining control of the execution
and collecting memory snapshots using the Java Virtual Machine Tool
Interface (JVMTI).  For more details on sampling and tracing we refer
to~\prettyref{appendix:tracing}.

Once tracing is completed, this data is used to train the neural ranking
function where PyTorch~\cite{NEURIPS2019_9015} is used as the machine learning
framework.  Finally, we encode the problem of certifying the neural ranking
function into an SMT formula and use
Z3~\cite{DBLP:conf/tacas/MouraB08} to solve it. More details about the
implementation of this step can be found in Appendix~\ref{sec:verification}.

\subsection{Benchmarks and Setup}
We consider three sets of programs for our experimental evaluation.  The
first and second set are comprised of problems from the TermComp Termination
Competition~\cite{DBLP:conf/tacas/GieslRSWY19} and the the SV-COMP Software
Verification Competition~\cite{10.1007/978-3-030-45237-7_21}.  Both problem
sets are publicly available and cover a wide variety of termination and
non-termination problems as well as software verification in general.

From these sets, we discard the non-terminating programs as they do not have
ranking functions.  Furthermore, for this evaluation we consider deterministic
programs with a maximum of two nested loops without function calls.  Hence, we
focus on the two problem sets \texttt{Aprove\_09} from TermComp and
\texttt{term-crafted} from SV-COMP.  After dropping problems due to the
aforementioned constraints we are left with 72 problems from
\texttt{term-crafted} (originally 159) and 38 problems (originally 76) from
\texttt{Aprove\_09}.  The problem set \texttt{term-crafted} comprises problems
from literature on termination analysis, which are given as C code.  We
therefore translate these problems into Java by hand.  In addition, we
also split the main functions' bodies into an initialisation and loop part.
Note that this purely syntactic change does not alter the difficulty of
determining termination of a program.  Finally, we present an additional problem
set -- {\texttt{nuTerm-advantage} -- consisting of problems created by us for
  showcasing notable strengths and weaknesses of the tested tools.

  To answer \emph{RQ2}, we compare \ourTool{} to
  \texttt{Ultimate}~\cite{HeizmannHP14},
  \texttt{Aprove}~\cite{GieslABEFFHOPSS17}, and
  \texttt{DynamiTe}~\cite{LeAFKN20}, which, collectively, represent the state of
  the art in termination analysis.

\paragraph{Setup} The experiments were conducted on Linux Kernel 5.15
running on an Intel Core i7 5820K at 3.3\,GHz with 16\,GB RAM and an NVIDIA GTX 980 graphics
card. For learning we use the Adam optimiser provided by PyTorch and a learning rate
of $0.05$. We run the benchmarks 5 times with random seeds that were fixed
a priori for reproducibility. Full instructions on how to reproduce the
results (including the seeds) are part of the supplementary material. We ran all
tools with a timeout of 60 seconds for each problem.

\begin{comment} \centering
  \includegraphics[scale=0.4]{timings}
  \caption{Percentage of problems solved over number of iterations required
    (accumulative) and percentage of problems solved within a given time spent
    on learning(in seconds).}\label{fig:timings}
\end{comment}
%
%
\subsection{Experimental Results}

\paragraph{Can neural ranking functions be used to formally prove the
  termination of software programs?}
To answer this question we ran \ourTool{} on the three benchmark sets
mentioned above.  The results are given in \prettyref{tab:results}.  We
observe the best performance of \ourTool{} when using a neural network
consisting of 7 neurons with $1000$ sample traces with a maximum length of
$1000$.  The exact strategy used is described in the supplementary material. 
\ourTool{} proves termination for 97.2 out of 124 problems on average (100
in the best out of the five runs), which accounts for 78.4\% of the problems
in the problem set.  When considering the different problem sets separately
we solve 91.0\% (\texttt{Aprove\_09}), 68.9\% (\texttt{term-crafted}), and
92.9\% (\texttt{nuTerm\_advantage}) of the problems.  Note that even when
disregarding the \texttt{nuTerm\_advantage} set, \ourTool{} solves 76.6\% of
the problems.  Given that very simple neural networks suffice to prove
termination of a substantial subset of the standard benchmarks, we answer
\emph{RQ1} in the affirmative.
\paragraph{Do neural ranking functions advance the state of the art in
termination analysis?}
To compare with the state of the art, we also ran
\texttt{Ultimate}~\cite{HeizmannHP14}, \texttt{AProVE}~\cite{GieslABEFFHOPSS17},
and \texttt{DynamiTe}~\cite{LeAFKN20} on the same benchmarks. Since
\texttt{Ultimate}~\cite{HeizmannHP14} and \texttt{DynamiTe}~\cite{LeAFKN20} do
not support Java code as input we ran the experiments on the C versions of the
problems. The results are presented in \prettyref{tab:results}. Overall, the
strongest tool is \texttt{AProVE}, which solves 81.4\% of all problems, followed by
\ourTool{} with 78.4\% and \texttt{Ultimate} with 77.4\% and finally \texttt{DynamiTe} with
67.7\%.  By considering the preexisting data sets separately, we see that
\ourTool{} comes in first on the \texttt{Aprove\_09} set and third on
\texttt{term-crafted}. On these two sets combined, \ourTool{} solves 76.6\% of
the problems with \texttt{AProVE} and \texttt{Ultimate} solving 89.1\% and
86.4\% respectively and \texttt{DynamiTe} trailing with 70\% of problems
solved. We conclude that on the existing benchmarks, \ourTool{} performs
comparably to the state of the art.

Our hypothesis is that \ourTool{} advances the state of the art when applied
to programs that have either disjunctive loop conditions or programs that
are nonlinear.  However, the existing benchmark sets suffer from
confirmation bias, and focus on programs that avoid these features.  To show
that \ourTool{} indeed advances the state of the art, we have compiled the
\texttt{nuTerm-advantage} data set, with programs that feature
\begin{enumerate}
\item non linear conditions, and\label{enum:linear-cond}
\item disjunctions in conditions.\label{enum:djsj-cond}
\end{enumerate}
The following code snippet is from
\verb|DynamiteExampleX4.java| from the \texttt{nuTerm-advantage} problem set:
\begin{minted}[escapeinside=||]{Java}
int a = 0;
while ( a*a*4 <= n ) {
  a = a + 1;
}
\end{minted}
This loop only uses two variables, $a$ and $n$, where $n$ remains constant
throughout the execution while $a$ is incremented by 1 in every iteration. 
Despite the fact that the loop guard $a^2\cdot 4 \leq n$ is nonlinear, there is a
linear ranking function.  Furthermore, the execution traces of the loop only
show the incrementing of $a$, which is also linear.  Our tool \ourTool{} can
solve this problem with a tiny neural network, consisting of a single
neuron, and reports the ranking function $\relu(n - a + 1)$.  Neither
\texttt{Aprove\_09} nor \texttt{Ultimate} are able to prove termination of
this problem, but \texttt{DynamiTe}, a tool which also utilises execution
traces, can solve it.
 
Disjunctions increase the complexity of formal reasoning significantly. 
This is illustrated by the following code snippet from
\verb|Square2VarsDisj.java| in the \texttt{nuTerm-advantage} set:
\begin{minted}[escapeinside=$$]{Java}
int a = 0, b = 0;
while ( a*a <= m || b*b <= n ) {
  a = a + 1;
  b = b + 1;
}
\end{minted}
None of the tools we compare with can show termination of this loop,
while \ourTool{} proves termination by learning the ranking function
$\relu(m-a + 2) + \relu(n-b + 2)$.  Note that \texttt{DynamiTe} is able to
show termination if the loop head was either~%
\mintinline[escapeinside=||]{python}{while |$a^2$| |$\leq$| m} or %
\mintinline[escapeinside=||]{python}{while |$b^2$| |$\leq$| n}. %
However, once both conditions are connected with a disjunction,
\texttt{DynamiTe} fails to show termination.

The \texttt{nuTerm-advantage} problem set comprises problems that exhibit
either nonlinear conditions, disjunctions, or a combination of both.  For
this set, \ourTool{} solves on average 92.9\% of the problems while
\texttt{DynamiTe} comes second, solving 50\%, followed by \texttt{AProVE}
(21.4\%) and \texttt{Ultimate} (7.8\%).

In conclusion, the experiments conducted and presented in
\prettyref{tab:results} show that on existing benchmarks, \ourTool{} either
performs either comparable to or stronger than (e.g., on \texttt{Aprove\_09}) the
state of the art.  Furthermore, we identified weaknesses in the existing tools
when considering a broader range of programs and show that neural termination
analysis can solve these problems by providing a set of programs on which
\ourTool{} outperforms all existing tools by a large margin.  Thus, we can answer
\emph{RQ2} in the affirmative.

\paragraph{How do neural ranking functions scale in terms of the complexity of
  the program?}
Our experiments have only required tiny neural networks, consisting of no
more than 10 neurons.  The results presented in \prettyref{tab:results} were
obtained by a neural network consisting of 7 neurons.  Increasing the number
of neurons further does not yield significant gains on the existing problem
sets as shown in \prettyref{fig:nodes-results}. We hypothesise that programmers avoid
writing loops that require termination arguments that depend on a very large
number of variables.
\begin{figure}
  \includegraphics[scale=0.58]{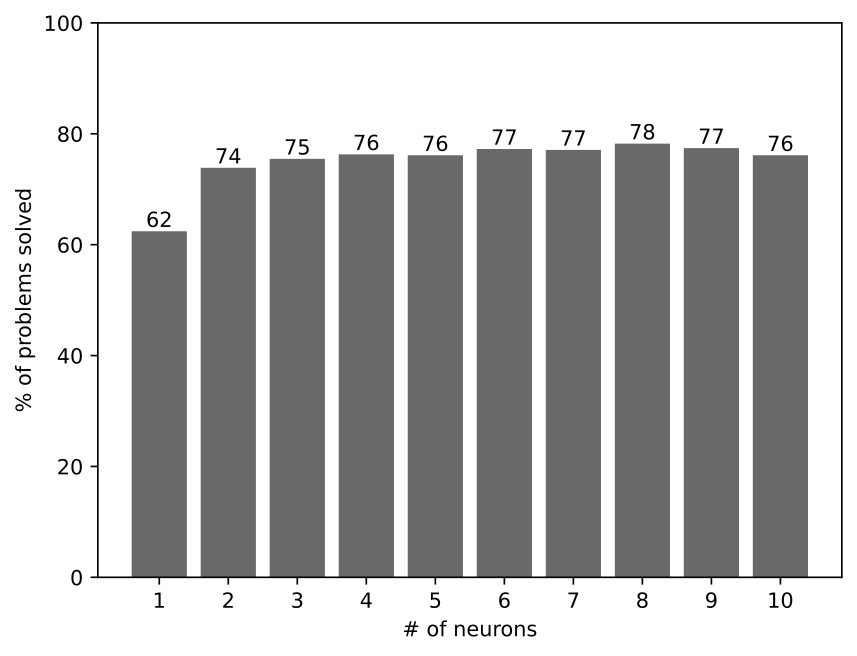}
  \caption{Percentage of problems solved for neural networks with 1
    to 10 hidden neurons.}\label{fig:nodes-results}
\end{figure}
To evaluate how our technique scales in the number of variables that are
required for the ranking argument, we use the following program template,
designed to require at least $k$ neurons.
\begin{minted}[]{Java}
int a = 0;
while ( a*a*4 <= n_1 || ... || a*a*4 <= n_k ) {
  a++;
}
\end{minted}
We include the instances of this template for values of $k$ up to $4$ in the
\texttt{nuTerm-advantage} problem set. Note that neither \texttt{AProVE} nor
\texttt{Ultimate} is able to solve this problem for any value of $k$;
\texttt{DynamiTe} is able to solve these problems for $k$ up to $2$ (with a
noticeable increase in runtime from 8\,s to 30\,s), but times out for any larger
$k$. Note that \ourTool{} solves these problems in the problem set within 2
seconds. Trying programs with $k$ up to 10 it becomes clear that
the learning procedure continues to scale well while the verification starts to become
the bottleneck.

It is worth emphasising that neural networks with $7$ neurons, for example,
which would be able to prove such a loop for $k = 7$ terminating, are
laughably small compared to state-of-the-art neural networks used in other
areas such as natural language processing.  Hence, it is likely that by
increasing the size of the neural network one would sooner run into issues
verifying the neural networks and generating meaningful traces than training
the neural networks.  Furthermore, we hypothesise that the vast majority of
loops in real world programs do not have termination conditions that involve
hundreds of disjuncts.  In conclusion, we can tentatively answer \emph{RQ3}
by suggesting that neural termination analysis scales well in the complexity
of the program, noting that other parts such as verification and tracing
might not scale as well.

\subsection{Discussion}

Owing to the inherent difficulty of the problem at hand (termination),
methods for solving it are necessarily incomplete.  Neural termination
analysis is no exception.  Under this constraint we presented an
experimental evaluation to answer three research questions regarding the
efficacy of neural ranking functions (\emph{RQ1}), their advantages over
other approaches (\emph{RQ2}), and their scalability (\emph{RQ3}).  Despite
the favourable answers for each of the questions it is important to point out
weaknesses of our approach.  For example, it is easy construct programs that
other methods can prove terminating but where neural termination analysis
fails.  Usually, these fall into one of the following three cases.

\paragraph{Insufficient Data}
Neural termination analysis learns termination arguments from execution
traces. Hence, any program feature that limits the data that can be
collected is a problem for our approach. Several benchmarks in the dataset
exhibit behaviour such as the following:
\begin{minted}{Java}
while (x > 0) {
  x = -2*x + 10;
}
\end{minted}
This loop has more than one iteration if and only if $x$ is one of $1,2,3,4$ and
even then the trace is extremely short. Similar issues can occur when offsets or
certain program branches only occur in rare cases. Such instances may lead to
overfitting of the networks to the sampled traces. As a result, a learned neural
ranking function may satisfy all required properties over the sampled traces but
not when verifying it with respect to all possible inputs in the verification
procedure.  Solver-driven test input generation may be a means to ensure that
traces for these behaviours are included in the training data set. Another issue
related to insufficient data is the existence of large constants in the
problems. For instance, considering a problem such as the following:
\begin{minted}{Java}
while (x < 10000000) {
  x++;
}
\end{minted}
This would require complete traces (letting $x$ go all the way to
$10000000$), which take a long time to gather.  Furthermore, learning a bias
that large can also pose problems.
\paragraph{Model Expressivity}
As ranking functions become more complex, we need neural architectures that
are able to express them.  One instance from the dataset where this problem
manifests is a benchmark where the ranking function depends on whether an
input variable is even or odd.  None of the neural architectures discussed
in Sec.~\ref{sec:learning-ranking-functions} is expressive enough to capture
the concept of ``even'' and ``odd''.  One way of solving this problem is by
considering further neural architectures, which would require a more
sophisticated data collection.  The key limiting factor when deploying such
architectures will likely be the increased complexity of the verification
process, rather than the learning.

\paragraph{Verification}
When there are multiple correct ranking functions the verification procedure
may not be able to prove all of them correct. The following loop can exhibit
such behaviour:
\begin{minted}{java}
int j = i;
while (i < 100) {
  i++;
  j = i;
}
\end{minted}
When purely looking at the execution traces, which is what the learning
procedure does, $i$ and $j$ have the exact same values at the loop head. 
Hence, if the learning process comes up with the ranking function $100 - j$
the verifier would not be able to prove it correct unless it is supplied
with the auxiliary invariant that $i = j$ at the loop head.  One way to
solve this problem could be to integrate existing methods that discover such
invariants~\cite{SiDRNS18, DBLP:journals/scp/ErnstPGMPTX07}.

\section{Threats to Validity}

We discuss threats to the validity of our experimental claims.

\paragraph{Benchmark Bias}
All our claims depend on the choice of benchmark programs. We focus on
sequential Java programs, and programs in other programming languages, or
programs that use concurrency, may require ranking functions that our neural
networks cannot find. While we use standard benchmarks from the
literature introduced by others to enable a comparison of different termination tools,
these benchmarks may not be representative for software written by developers.
Moreover, a source of bias may be introduced by our collection of programs
with either disjunctive conditions or nonlinear behaviour.
While we believe that both features are
important in commodity software, it remains to be quantified how large the
benefit of supporting these features is on a larger repository of software.

\paragraph{Test input generation}
We require that test inputs can be generated that exercise the programs to
yield traces to train the neural network.  While our simple sampling
method is successful on our benchmarks, it may not be possible in general to
obtain sufficiently diverse test inputs.  More sophisticated means to
generate test inputs are known, and can be used to mitigate this threat.

\paragraph{Neural architecture complexity}
While our experiments suggest that tiny neural networks are able to prove
termination of most program loops, there may exist programs that require a
large number of neurons, increasing training and verification complexity.

\paragraph{Auxiliary Invariants}

We use a standard verification step for checking the validity of the neural
ranking function, which, in some cases, requires an auxiliary invariant. 
The results of an end-to-end termination analysis are dependent on the
quality of these invariants, and the tools we compare with use a variety of
different approaches to solve this problem.  Our own tool uses a simplistic
heuristic for guessing these invariants (Appendix~\ref{sec:verification}),
which may be exceptionally successful.  The documentation on the algorithms
for generating these invariants is limited, and a proper comparison of
alternative methods for generating ranking functions requires an
implementation in a single framework.

\section{Related Work}
\subsection{Termination Analysis}

Many methods for automatically proving termination have been
developed and implemented. Owing to
the undecidability of the problem in general, most
techniques restrict the scope
of the analysis in some way, e.g., to linear ranking functions and
programs~\cite{PodelskiR04,DBLP:journals/jacm/Ben-AmramG14,DBLP:conf/pldi/GonnordMR15},
semi-definite programs~\cite{Cousot05}, semi-algebraic
systems~\cite{ChenXYZZ07}, or formulae drawn from specific SMT
theories~\cite{DBLP:journals/fmsd/CookKRW13}. To deal with complex program
loops, lexicographic ranking functions~\cite{BradleyMS05, icalp/BradleyMS05,
  10.1007/978-3-642-54862-8_12},
piecewise ranking functions~\cite{Urban13,DBLP:conf/tacas/Urban15},
disjunctively well-founded transition invariants~\cite{DBLP:conf/lics/PodelskiR04,pldi/CookPR06,
  10.1007/978-3-642-36742-7_4, DBLP:conf/cav/KroeningSTW10}, and implicit
ranking functions have been used~\cite{DBLP:conf/kbse/ChenH20}.
Ranking functions have been synthesised symbolically using, e.g., Farkas' lemma
and template-based guess-and-check
strategies~\cite{UrbanGK16,DBLP:conf/cav/FedyukovichZG18}.
To prove conditional termination, also abstract interpretation (by underapproximation)
and loop summarisation methods have been used~\cite{DBLP:conf/cav/CookGLRS08,DBLP:conf/tacas/TsitovichSWK11,DBLP:conf/popl/CousotC12,DBLP:journals/toplas/ChenDKSW18,DBLP:conf/pldi/0001K21}. 

Alternative methods perform termination analysis by translating programs to
alternative models of computation and show that the resulting model is terminating. This requires a guarantee that termination of
the translated program implies the termination of the original program. 
Models used for this purpose include
term rewriting systems~\cite{GieslABEFFHOPSS17,BrockschmidtOEG10, rta/OttoBEG10}, constraint logic programs~\cite{DBLP:conf/sas/Spoto16}, recurrence
relations~\cite{DBLP:conf/fmco/AlbertAGPZ07}, and Büchi
automata~\cite{HeizmannHP14,DBLP:conf/pldi/ChenHLLTTZ18}.

More recently, SMT solving has been used to discover ranking functions
from execution traces~\cite{LeAFKN20}, similarly to methods based on machine learning.

\subsection{Machine Learning for Termination Analysis}

In the last years, several termination analysis approaches that
incorporate machine learning technologies have been presented.
Early methods learn linear ranking functions
from execution traces by constructing a linear regression problem
whose solutions describe a loop bound~\cite{sigsoft/Nori013}.
Recently, machine learning models such as Support Vector
Machines (SVM) have been used as representation for
ranking functions in~\cite{DBLP:journals/access/YuanL19}.
Methods based on SVM have been applied to 
single or nested loop programs defined using conjunctions of
continuous functions for the guard, and deterministic
assignments defined as continuous functions as well~\cite{DBLP:conf/icfem/LiS0T019}. 

Another deep learning approach for termination analysis
has recently been introduced~\cite{DBLP:journals/nca/TanL21}.
This method uses neural networks with sigmoidal
activation functions, which are shown to be an appropriate ranking
function representation for programs defined using continuous functions,
without disjunctions and conditional choices.
While this is suitable to describe deterministic dynamical systems
in discrete time, this language
restriction makes the method inapplicable to software,
including the majority of our simple termination analysis benchmarks.
We estimate that 46 out of 110 (cf.\ combined $\{1,2\}$ in \prettyref{tab:results})
programs in the preexisting benchmark sets are in
the scope of (but not necessarily solved by) their method and remark
that our method solves 39 out of these 46 problems.
Besides, 5 out of the 14 benchmarks in \texttt{nuTerm\_advantage}  ($\{3\}$) are in
their scope, all of which are solved by \ourTool{}.
Unfortunately, we cannot directly evaluate the effectiveness of their method on
our benchmark set (neither $\{1,2\}$ nor $\{3\}$), because an implementation is unavailable.
Moreover, their method cannot be easily implemented in our infrastructure.
In fact, neural ranking functions with sigmoidal activation lack
efficient---and complete---decision procedures for checking their validity.
Notably, their approach required the development of bespoke decision procedures for this purpose.
Conversely, our method uses ReLU activation functions,
which can be encoded into expressions in decidable theories,
for which efficient SMT solvers are available. 
Our work goes a step further by showing that neural networks with ReLU activation
functions are sufficient to obtain results that are comparable to
state-of-the-art tools and even enable the effective termination analysis of
programs that are beyond their reach.

Recently, a data-driven method has taken
a similar approach and employed efficiently checkable
templates to learn loop bounds~\cite{boundLearningForTermination}.
They propose a portfolio of methods and templates
for this purpose. By contrast, our method employs neural
networks whose expressive power subsumes a wide variety
of ranking function templates. Our learning phase
only relies on optimising a loss function, and can thus
be implemented using generic optimisation
algorithms that are readily available
in machine learning frameworks.

A heuristic approach to termination analysis based on deep
learning has been proposed
by Alon and David~\cite{DBLP:journals/corr/abs-2207-14648}.
This approach uses graph neural networks on a program's
abstract syntax tree to estimate the likelihood of
termination. Specifically, this results in attention
networks that propose locations in the program that might be a cause for
non-termination. Importantly however, the proposed method trains 
neural networks over a large dataset of terminating
and non-terminating programs in a supervised learning fashion.
The approach is envisioned to be part of a debugging workflow
where potential issues are
highlighted for consumption by a programmer or another analyser,  
and it is not aimed at providing formal proofs of termination;
therefore, it does not give correctness guarantees.
By contrast, our approach has a distinct training phase for each
program and does not rely on any a priori knowledge. Moreover,  
it provides a formal certificate of
termination---the neural ranking function---whose
validity we check using SMT solving.
Our method is thus fully unsupervised and provides formal
guarantees of termination
when a valid neural ranking function is found. 

\subsection{Deep Learning for Automated Reasoning}
Neural networks have been used in other areas of automated reasoning and verification.
In automated and interactive theorem proving,
neural networks have been used for premise
selection~\cite{DBLP:conf/nips/IrvingSAECU16} and proof search in first and
higher-order logic~\cite{DBLP:conf/ecai/OlsakKU20,DBLP:conf/aaai/PaliwalLRBS20}. 
Recent approaches have made use of deep learning in
program and control synthesis~\cite{DBLP:conf/iclr/ParisottoMS0ZK17,DBLP:conf/corl/DawsonQGF21,DBLP:conf/corl/SunJF20,DBLP:conf/iclr/QinZCCF21}.
In software verification, deep learning has been used to find 
loop invariants~\cite{SiDRNS18,DBLP:conf/sas/BrockschmidtCKK17,DBLP:conf/iclr/RyanWYGJ20}.

Our method falls within the realm of approaches that use
neural networks {\em to represent}, rather than {\em to output},
formal certificates of correctness with soundness guarantees.
Exemplars are Lyapunov neural networks and their
generalisation into neural barrier certificates, which
have been used for the formal stability and safety analysis
of dynamical systems~\cite{ChangRG19,AbateAGP21,DBLP:conf/hybrid/ZhaoZC020,DBLP:conf/tacas/PeruffoAA21,DBLP:conf/hybrid/AbateAEGP21,DBLP:conf/rss/DaiLYPT21,DBLP:conf/hybrid/ChenFMPP21}.
More recently, the neural ranking supermartingale model has been
introduced for the termination analysis of probabilistic programs,
and subsequently applied to the stability analysis
of stochastic control systems~\cite{DBLP:conf/cav/AbateGR20,DBLP:conf/aaai/LechnerZCH22}. In a similar fashion, also decision tree learning
has been applied to the verification
of probabilistic programs~\cite{DBLP:conf/cav/BaoTPHR22}.

%queries~\cite{DBLP:conf/icml/FischerBDGZV19},

\subsection{Formal Verification of Neural Networks} Automated reasoning
methods that use neural networks as representation of proof certificates,
including our approach, rely on formal verification technologies to check
the validity of these neural certificates.  Various methods for the formal
verification of neural networks have been developed in the last few years,
driven by the quest for formal guarantees against adversarial attacks in
computer vision~\cite{DBLP:journals/corr/SzegedyZSBEGF13}.  Significant
effort has been made towards this goal by using out-of-the-box SMT
solvers~\cite{DBLP:journals/aicom/PulinaT12} and, subsequently, developing
tailored methods to reason about neural networks.  This effort led to the
development of many effective tools and
algorithms~\cite{DBLP:journals/ftopt/LiuALSBK21,
DBLP:conf/cav/HuangKWW17,DBLP:conf/atva/Ehlers17,
DBLP:conf/nips/SinghGMPV18, DBLP:conf/cav/KatzBDJK17,
DBLP:conf/nips/ZhangWCHD18,DBLP:conf/cav/KatzHIJLLSTWZDK19,
DBLP:conf/ecai/HenriksenL20,DBLP:conf/icse/ShriverED21,
DBLP:conf/nips/BunelTTKM18,
DBLP:conf/ijcai/KouvarosL21,DBLP:conf/cav/TranBXJ20}. 

Methods for adversarial attacks reason about neural networks in isolation,
while reasoning about neural ranking functions requires reasoning about
neural networks together with (1) constraints arising from the encoding of a
program which (2) are usually in theories other than real arithmetic such as
integer or bit-vector arithmetic.  Reasoning about neural networks in
combination with other systems has been treated in the context of safety
analysis of neural network controllers for dynamical
systems~\cite{DBLP:conf/hybrid/IvanovWAPL19, DBLP:conf/cav/TranYLMNXBJ20,
DBLP:conf/icra/VincentS21,DBLP:conf/ijcai/BacciG021,
DBLP:conf/aaai/0001FG22}.  Methods of this kind apply abstract
interpretation for a bounded number of steps or compute invariants. 
Verification of neural networks under different theories has been considered
for binarized and quantized neural networks, in isolation from other systems,
specifically for adversarial attack
problems~\cite{DBLP:conf/aaai/NarodytskaKRSW18,
DBLP:conf/tacas/GiacobbeHL20, DBLP:conf/aaai/HenzingerLZ21,
DBLP:conf/tacas/Amir0BK21}.

In our work, we use off-the-shelf SMT solving to check neural ranking
functions because, in our experiments, we rely on relatively small networks. 
While we observe that for a wide variety of problems small networks are
sufficient, we do not preclude that our method may benefit from using larger
networks.  Using larger networks may pose limits to the scalability of
off-the-shelf SMT solvers.  Our work adds a novel problem to the spectrum of
formal verification questions for neural networks, contributing to their
relevance to software engineering applications beyond robustness to
adversarial attacks.

\section{Conclusion}\label{sec:conclusion}

We introduced a termination analysis method that takes advantage of neural
networks by learning a ranking function candidate from sampled execution
traces.  This neural ranking function is subsequently verified using formal
methods.  We provide a prototype implementation of this method called
\ourTool{}.  When using tiny neural networks with one hidden layer and a
straight-forward training script we solved 76.6\% of the problems
in a standard set of benchmarks for termination analysis performing
comparably to state-of-the-art tools.  Furthermore, we identify problems
with disjunctive or non-linear loop guards where competing tools are unable
to prove termination.  We show, experimentally, that \ourTool{} is able to
solve these types of problems by creating a separate set of benchmarks that
feature such loop guards.  On this set, \ourTool{} can solve 94.3\% of the
problems, outperforming competing tools.

Our result suggests future research both in machine learning and formal
verification.  Learning proof certificates from examples applies not only to imperative
programs, but also to functional programming and logic.  Moreover,
separating proof learning from formal proof checking may also apply to
further verification tasks such as non-termination~\cite{GuptaHMRX08}, which
is difficult for conventional formal approaches.

\appendix
\section{Tracing}\label{appendix:tracing}
A trace is a sequence of snapshots (observations) of the program's state as the program runs.
Traces are thus generated dynamically, by running the program.
The process of tracing takes two inputs: the program that is to be traced and a list of
program locations in the program where a snapshot of the state is to be
taken. In our case these locations are the loop heads in the program.  Our
approach is conceptually simple and independent of the platform and
programming language. As we consider Java, we give implementation
details specific to the Java environment and the Java Virtual Machine (JVM),
but our approach could also be applied to more abstract models of
computation.
 
Tracing consists of three steps:
\emph{input sampling}, \emph{execution}, and \emph{snapshot}.  We describe
each of these steps below.

\paragraph{Input sampling}
Termination analysis is commonly applied to program fragments that contain
some initialisation and a (possibly nested) loop. Therefore, we work with programs
that are not closed, but require inputs.  We only consider
deterministic programs, i.e., two traces that are generated with the same
sequence of inputs are identical.  We use
two sampling strategies based on a Gaussian distribution: \emph{pairwise
anticorrelated sampling} (PAS) and \emph{Gaussian sampling}.
PAS uses a multivariate normal distribution where we enforce the same
variance for all inputs except for two randomly chosen inputs.  For these
two chosen variables we create a covariance.  Hence, this is a standard
Gaussian distribution for all variables except for two where we enforce a
covariance.
Gaussian sampling, on the other hand, is a sampling strategy where each
variable is sampled independently from another from a Gaussian distribution
with a variance of $1000$ and no covariance.

\paragraph{Execution}

We start executing the program with the sampled arguments.  We maintain
control over the JVM during the execution using the Java Virtual Machine
Tool Interface (JVMTI).  Once we hit a loop head location, we halt the
execution and take a snapshot.

\paragraph{Snapshot}

Using the JVMTI we have access to the Local Variable Table (LVT).  The LVT
contains all local variables of the function.  We create a memory snapshot
by iterating through the LVT and reading the values of every variable that
is in scope at the given location.  For variables that are out of scope, we
record a placeholder default value (which depends on the type of the
variable).  Since the number of local variables does not change, the size of
the snapshots is always the same.  Once the snapshot is collected, we append
it to the trace of the current program.  If the maximum length of a trace is
reached we force a termination of the virtual machine, otherwise we resume
the execution.
The resulting list of snapshots constitutes an execution trace.  The goal is
to sample the input data in such a way that we achieve a high coverage of
the function and the data therefore best represent a possible ranking
argument while keeping the required number of program runs low.
Our experiments with different sampling strategies show that PAS exhibits better
performance than multivariate gaussian sampling.  When using the same neural
network with 10 neurons we solve 79\% of problem using traces obtained
with PAS and 74.8\% when using multivariate gaussian sampling.  It should be
noted that both sampling techniques are extremely simplistic.  The results
may be further improved by utilising more sophisticated test input
generation or fuzzing~\cite{DBLP:conf/issta/VisserPK04,
DBLP:journals/tse/BaludaDP16, DBLP:journals/jss/ChenKMT10,
10.1007/978-3-540-30502-6_23, DBLP:conf/icse/PachecoLEB07,
DBLP:journals/cacm/CadarS13, cispa3120}.

\section{Verification}\label{sec:verification}\label{appendix:verification}

We verify that a candidate (monolithic) neural raking
function $f \colon \Theta \times \Real^n \to \Real$
is a valid neural ranking function for program $P = (S,T)$ by verifying
that it decreases every time a
loop header is encountered. For this purpose, we construct a symbolic encoding of the
transition relation between every two loop headers $T_H$.
Thus the verification question corresponds to that of determining whether,
for any two states between loop headers that also satisfy an auxiliary invariant,
the trained neural ranking function decreases by $\delta > 0$. This corresponds to checking
the following validity question:
\begin{equation}
  \forall s,s' \colon (s,s') \in T_H \land s \in A \implies
  f(\theta;\omega(s)) \geq f(\theta;\omega(s')) + \delta \label{eq:verif}
\end{equation}
Note that $\theta$ is constant is this formula.
Then, if this formula is valid, we have that $f(\theta ;\,\cdot\,)$ is a valid ranking function.
We verify this by checking the dual satisfiability question using an SMT solver.
The dual satisfiability question is that of finding a counterexample where the candidate does not decrease by $\delta$,
that is, the following formula:
\begin{equation}
  \exists s,s' \colon \underbrace{(s,s') \in T_H \land s \in A \land
    f(\theta;\omega(s)) < f(\theta;\omega(s')) + \delta}_\varphi 
\end{equation}
If the quantifier-free formula $\varphi$ is determined unsatisfiable by the SMT
solver, then the ranking function is valid.

Encoding $\varphi$ involves encoding constraints for the program $T_H$ and $A$,
constraints for the neural ranking function $f$, and the interface between
them which is the observation function $\omega$.
We encode the transition relation between loop headers $T_H$ using a 
single static assignment encoding, which introduces intermediate variables
after each assignment and encodes operations using appropriate
arithmetic expressions. This is similar to a bounded model checking
encoding~\cite{DBLP:conf/tacas/ClarkeKL04,DBLP:conf/cav/CordeiroKKST18},
which is possible because every run between adjacent loop headers has necessarily fixed length.
According to the semantics one wants to consider, the program can be encoded in the theory
of integers or the theory of bit-vectors (in our experiments, we use the theory of integers).
Also $f$ and $\omega$ can be seen as bounded programs
(note that $f$ is a feed-forward network)
and therefore can be encoded similarly.
Our neural networks are compositions of linear layers and ReLU activation functions,
which result in first-order logic formulae in the theory of reals with linear arithmetic. 
Notably, we use much smaller networks compared to common machine learning
applications, and we argue that these 
are sufficient to solve a broad variety of termination problems.
Also our program encoding is smaller than those generated by
bounded model checkers, as it involves at most one loop unrolling.
For this reason, our overall encoding ultimately results in
formulae that are efficiently solvable by modern SMT solvers. 

Identifying auxiliary invariants $A$ that are strong enough for termination analysis
is a difficult problem (encoding them is straightforward).
Our verification procedure uses a heuristic that syntactically extracts constraints from the program from,
for example, conditional statements, and checks whether these are loop invariants using the SMT solver.
If these are valid invariants, then they used to define $A$. As it turns out, this naive heuristic was
sufficient to obtain the results presented in this paper. Using more sophisticated loop invariant generation
methods can only improve the effectiveness of our tool and is subject of future investigation.
Methods for generating loop invariants include procedure based on theorem 
provers~\cite{DBLP:conf/fase/KovacsV09},
constraint based invariant synthesis~\cite{DBLP:conf/vmcai/BeyerHMR07,DBLP:conf/cav/LahiriB04,DBLP:conf/pldi/BeyerHMR07}.
Invariants for Java Programs have been constructed using symbolic
execution~\cite{8115691}.  Tools for the discovery of invariants from
trace data include
Daikon~\cite{DBLP:journals/scp/ErnstPGMPTX07} and
DIG~\cite{DBLP:journals/tosem/NguyenKWF14}.

Similarly, for lexicographic neural ranking functions we verify the validity of
the conditions in Eq.~\eqref{eq:lex1} and \eqref{eq:lex2} over each
loop header and respective output component of the neural ranking function. 
Let $H = \{ h_1, \dots, h_m\}$ be the set of loop headers, then for
every $i = 1, \dots, m$ we verify the validity of the following conditions:
\begin{align}
  \forall s,s' \colon (s,s') \in T_{\{h_i\}} \land s \in A_i &\implies
  f_i(\theta;\omega(s)) \geq f_i(\theta;\omega(s')) + \delta \notag\\
  \forall s,s' \colon (s,s') \in T_{\{h_i\}} \land s \in A_i &\implies
  f_{i-1}(\theta;\omega(s)) \geq f_{i-1}(\theta;\omega(s'))\notag\\[-1.5ex]
  &\quad\vdots\notag\\[-0.5ex]
  \forall s,s' \colon (s,s') \in T_{\{h_i\}} \land s \in A_i &\implies
  f_{1}(\theta;\omega(s)) \geq f_{1}(\theta;\omega(s'))
\end{align}
Note that each condition can be checked independently: the lexicographic
argument is violated if any of the conditions is violated, otherwise it is valid.
We remark that unlike the monolithic case, $T_{\{h_i\}}$ may
represent runs of arbitrarily length when the loop with 
header $h_i$ has nested loops. To encode $T_{\{h_i\}}$ as a bounded problem
we substitute every inner loop with a summary.
Several methods have been developed for this purpose~\cite{DBLP:conf/atva/KroeningSTTW08,DBLP:conf/tacas/TsitovichSWK11,DBLP:conf/pldi/0001K21}.
As a heuristic, we construct a loop summary that encodes the invariance
of all variables that are never assigned within the loop, together with a transition
invariant that encodes the respective (and previously verified) lexicographic component in the
neural ranking function and the respective auxiliary invariant.
Using or developing more sophisticated summarisation techniques is matter of future research.

%\section{Acknowledgments}
\begin{acks}
  To appear in the proceedings of ESEC/FSE '22~\cite{fse22}.
  We are grateful to all anonymous reviewers
  for their helpful comments and suggestions. 
  This work was in part supported by the HICLASS project
  (\grantnum{}{113213}), a partnership between the \grantsponsor{}{Aerospace Technology
    Institute (ATI)}{}, \grantsponsor{}{Department for Business, Energy \& Industrial
    Strategy (BEIS)}{}, and \grantsponsor{}{Innovate UK}{}. This work
  was also in part funded by the \grantsponsor{}{Oxford-DeepMind}{} Graduate Scholarship as well as the
  \grantsponsor{}{Engineering and Physical Sciences Research Council (EPSRC)}{}. 
  For the purpose of Open Access, the authors have applied a CC
  BY public copyright licence to any Author Accepted Manuscript (AAM)
  version and the final version arising from this submission.
\end{acks}

%\clearpage
\bibliographystyle{ACM-Reference-Format}
\bibliography{neuraltermination}

\end{document}